\documentclass[final,5p,times,twocolumn]{elsarticle}

\usepackage{amssymb}
\usepackage[T1]{fontenc}
\usepackage{hyperref}
\usepackage{caption}
\usepackage{xcolor}
\usepackage{subcaption}
\usepackage{graphicx}
\usepackage{amsmath}
\usepackage{amsthm}
\newtheorem{definition}{Definition}
\usepackage{lineno}

\journal{Pattern Recognition}

\begin{document}

\begin{frontmatter}



\title{Convolutional Neural Networks Rarely Learn Shape for Semantic Segmentation}


\author[label1]{Yixin~Zhang}
\author[label1,label2,label3,label4]{Maciej~A.~Mazurowski}

\affiliation[label1]{organization={Department of Electrical and Computer Engineering, Duke University},
            city={Durham}, 
            state={NC},
            country={USA}}

\affiliation[label2]{organization={Department of Radiology, Duke University},
            city={Durham}, 
            state={NC},
            country={USA}}
\affiliation[label3]{organization={Department of       
Biostatistics and Bioinformatics, Duke University},
            city={Durham}, 
            state={NC},
            country={USA}}
\affiliation[label4]{organization={Department of       
Computer Science, Duke University},
            city={Durham}, 
            state={NC},
            country={USA}}
\begin{abstract}
Shape learning, or the ability to leverage shape information, could be a desirable property of convolutional neural networks (CNNs) when target objects have specific shapes. While some research on the topic is emerging, there is no systematic study to conclusively determine whether and under what circumstances CNNs learn shape. Here, we present such a study in the context of segmentation networks where shapes are particularly important. We define shape and propose a new behavioral metric to measure the extent to which a CNN utilizes shape information. We then execute a set of experiments with synthetic and real-world data to progressively uncover under which circumstances CNNs learn shape and what can be done to encourage such behavior. We conclude that (i) CNNs do not learn shape in typical settings but rather rely on other features available to identify the objects of interest, (ii) CNNs can learn shape, but only if the shape is the only feature available to identify the object, (iii) sufficiently large receptive field size relative to the size of target objects is necessary for shape learning; (iv) a limited set of augmentations can encourage shape learning; (v) learning shape is indeed useful in the presence of out-of-distribution data.

\end{abstract}

\begin{keyword}
Segmentation\sep Feature Measurement\sep Machine Learning\sep Computer Vision



\end{keyword}

\end{frontmatter}


\section{Introduction}
Since the breakthrough made by AlexNet on ImageNet, CNNs with different architectures have achieved promising results on image classification~\cite{shen2015robust,9174822,9598903}, segmentation~\cite{anand2023fusion,zhao2017pyramid,Chen_2018_ECCV}, and object detection~\cite{liu2022convnet} tasks. Early research attributed CNNs’ successes to the similarity between CNNs and a human’s visual system. It was presumed that CNNs learn simple low-level features (i.e., edges) in shallow layers and combine them to form complex global shapes until objects can be recognized~\cite{kriegeskorte2015deep,lecun2015deep}. 
Recent data suggest that CNNs utilize only limited shape information in a typical setting~\cite{geirhos2018imagenet,doi:10.1146/annurev-vision-120522-031739}. However, a systematic study of the issue is missing. If CNNs indeed do not learn shape, this property of CNNs is likely to lead to their limited ability to generalize in settings where noise distribution, brightness, or other appearance change.

Learning shape by CNNs could be of benefit. If we encourage CNNs to use shape information, they may learn and utilize a feature set more robust against common appearance disparities between training and testing images~\cite{mummadi2021does}. An example of an application area where this might be of particular benefit is medical imaging. For example, magnetic resonance imaging (MRI) and computed tomography (CT) images maintain the appearance of shapes for a given patient because they are typically filmed at a fixed position relative to body parts with similar 3D geometry. 
However, each image’s texture, noise distribution, and intensity/color can highly vary depending on a device's hardware specifications and imaging parameters. 
CNNs that learn shape could rely less on inconsistent characteristics but focus on the more invariant object shape. 

Shape-learning CNNs may also be more robust against adversarial attacks. Tremendous efforts and multiple algorithms have been proposed to counter malicious uses such as applying ensemble learning~\cite{DBLP:conf/iclr/TramerKPGBM18,shen2021bbas}, random smoothing~\cite{pmlr-v97-cohen19c}, adversarial training~\cite{shafahi2019adversarial}.  Recent research perceives adversarial vulnerability as a property of the data set rather than the network. Specifically, adversarial vulnerability arises from the presence of highly predictive but non-robust features uninterpretable to humans in the training set \cite{ilyas2019adversarial}. If we consider adversarial vulnerability arising from uninterpretable features,  CNNs trained to utilize human-interpretable shape features may show higher robustness against adversarial attacks. 

In this paper, we devise a systematic empirical study to analyze (1) whether CNNs can learn shape and (2) how to make them learn shape. Our study is conducted in the context of semantic segmentation, where shape plays a particularly important role.

Our contributions can be briefly summarized as follows:
\begin{itemize}
  \item We present the first systematic study on the role shape, texture, and structure play in the feature-learning of CNNs based on unambiguously defined concepts. These concepts have been often used without a precise meaning in previous literature.
  \item We propose a well-defined metric to quantify the behavior of a network and assess whether it considers shape during inference. Importantly, this metric is model-independent and thus can be applied to a variety of neural architectures.
  \item We devise an experimental framework with simulated as well as real-world data and we conduct extensive experiments to test whether and in what conditions CNNs learn shape. 
  \item We propose and evaluate data augmentation techniques to promote learning of shape in neural networks. 
\end{itemize}
These contributions will facilitate more rigorous and consistent thinking about shape and will inform researchers about the potential outcomes when choosing certain network architectures or performing data augmentations.

\section{Related Work}
A systematic study on how discriminative features in training data affect the shape learning of CNNs is lacking. However, there are some studies that provide limited, yet helpful information on specific aspects of this question.
Kubilius et al.~\cite{kubilius2016deep} created a validation set entirely of object silhouettes. They found that CNNs still showed predictive capability on the silhouettes, although their performance dropped by 30\%. Additionally, they found that objects in the same super-ordinate category have similar representations but offered little interpretation of this observation. 
Baker et al.~\cite{baker2018deep,baker2020local} show that texture plays a more significant role than global shape in ImageNet-trained CNN’s perception of objects, where CNNs barely learn the abstract relations of elements. The authors made no further comments on the cause of this phenomenon. 
Geirhos et al.~\cite{geirhos2018imagenet} also showed that ImageNet-trained CNNs rely heavily on textures for classification, instead of using a conflict-cue method. The authors suggest the potential of applying neural style transfer (NST) as an augmentation to encourage shape learning for better testing accuracy and OOD robustness.
Ritter et al.\cite{ritter2017cognitive} demonstrated that Inception and Matching Nets\cite{vinyals2016matching} pre-trained on ImageNet tend to generate latent representations with higher similarity for objects with similar shapes than those with similar colors. However, it would be premature to conclude that CNNs are shape-learning solely through their insensitivity to colors. 
Hermann et al.\cite{hermann2020origins,hermann2020shapes} show that CNNs with the same model architecture may have vastly different shape learning behaviors for different training datasets and objectives. Their configurations have direct impacts on the feature extraction process of the SN. Though innovative in its nature, their research ended with a conclusion that the “most decodable feature from untrained mode” will be learned, without a further account of what it practically means by “decodable”.
Hosseini et al.\cite{hosseini2018assessing} state that proper augmentation, initialization, and the use of batch normalization are essential for networks to learn shape. Li et al. \cite{li2020shape} combined training with stylized images, soft labeling, and weighted loss as a heuristic to gain more control over the inductive bias towards texture or shape during the model training. Their approach achieved improved performance on the ImageNet-C benchmark \cite{hendrycks2019benchmarking}. Islam et al. \cite{Islam2021ShapeOT} proposed a dimensionality estimation metric and used the metric to assess a CNN’s capacity to represent shape in latent space. Their study answers the questions “whether a CNN is capable of learning shape” and “where are those representations stored in the CNN” but did not address which factors encourage shape learning in a general setting. Other works also propose different approaches during model training to facilitate shape learning \cite{Tripathi_2023_CVPR,sinha2023towards,ding2022scaling} 

While the prior work has limitations, we learn the following. First, neural networks might not learn shape as often as one intuitively expected. Second, whether the network learns shape may depend on various factors, but it is not entirely clear which factors those are.

\section{Definitions}
\subsection{Terminology}
\label{def}
To enhance the clarity and rigor of our study, we state the definition of the terms commonly used in this paper. An \emph{object} in the context of images is a set of contiguous pixels. Three attributes of \emph{objects} are of particular interest here: \emph{shape, texture, and structure}. For any arbitrary \emph{object},
\begin{definition}
"Shape" is the spatial relationship between all boundaries separating pixels inside the object from pixels that are outside.
\end{definition}
The definition of \emph{texture} has been well-established in previous literature\cite{haralick1973textural,van1985texture}. We adopt it as:
\begin{definition}
"Texture" is a distribution from which pixels forming similar visual impressions may be sampled.
\end{definition}
Intuitively, each pixel can be associated with only one texture. We use the definition of texture to establish definitions of a \emph{simple object} and a \emph{complex object}, which will be of importance in this study. 
\begin{definition}
If all pixels within an object are associated with the same texture, this object is a "simple object"  Otherwise, this object is a "complex object" 
\end{definition}
\begin{definition}
"Structure" is the spatial relationships between the simple object(s) in each complex object. Simple objects all have the same "null" structure. 
\end{definition}
\subsection{Metrics}
\label{SBI}
Given these definitions of shape, texture, and structure, we can rigorously describe the meaning of “shape learning”. Conceptually, a shape-learning network should recognize objects with desired shapes even if the other features (e.g., textures) of the objects change. On the other hand, they ignore objects with irrelevant shapes but desirable non-shape features. To quantify the extent to which a Network network’s performance matches this behavior, we propose a metric named Shape Bias Index (SBI). This metric is named after the concept of Shape Bias in psychology \cite{diesendruck2003specific}, which is referred to as a model’s ability to recognize objects by their shape in computer vision. 
\begin{figure*}[t!]
\centering
\includegraphics[width=0.9\textwidth]{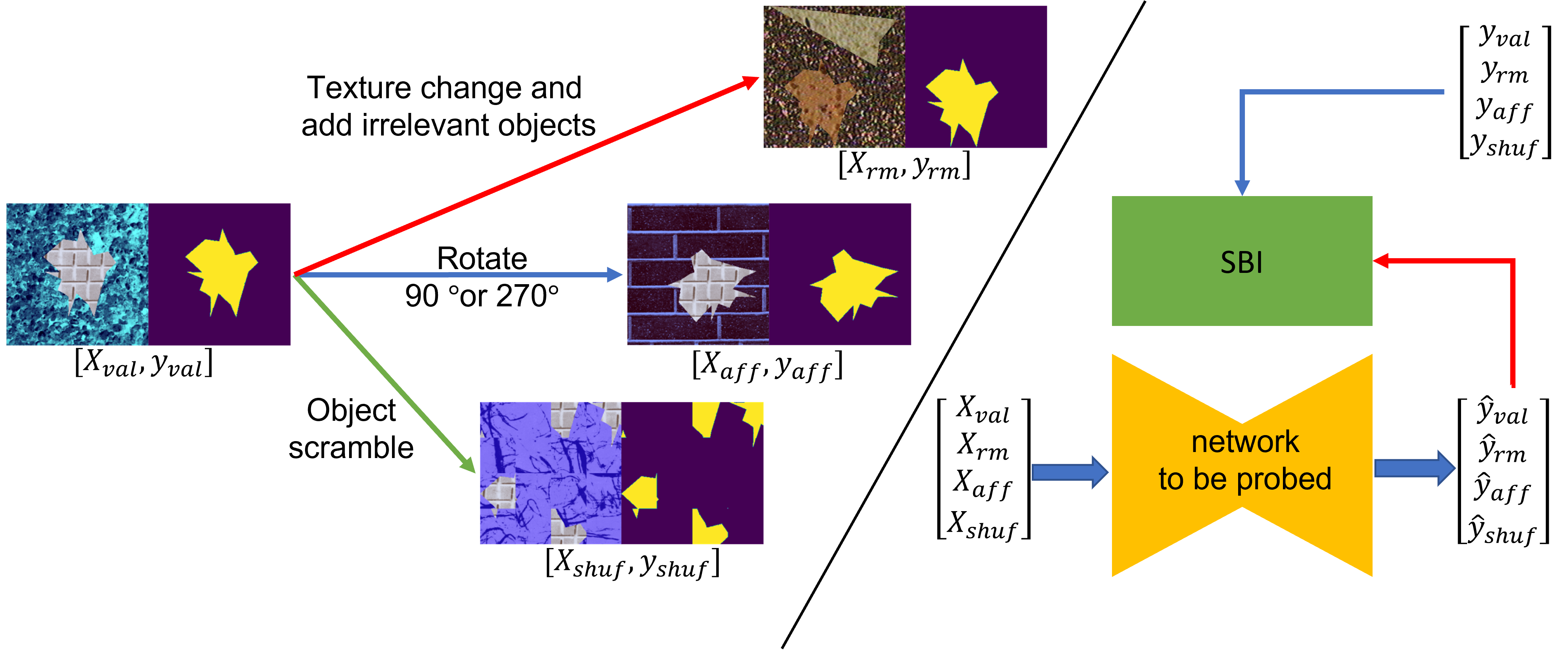}
\caption{\footnotesize Pipeline for (left) generating probing images and (right) computing SBI}
\vspace{-2mm}
\label{prob_img}
\end{figure*}

SBI is assessed empirically by isolating and comparing the effects of different discriminative features. Given a raw dataset, we split it to obtain a training set and a validation set such that the samples in the validation set contain a similar set of possible discriminative features to the training sets. We denote this validation set as $D_{val}=[X,y]$. Based on $D_{val}$, we then generate probing sets by removing certain discriminative features from $D_{val}$
\begin{itemize}
    \item $D_{rm}=[X_{rm},\ y_{rm}]$ by altering one or multiple presented non-shape discriminative features (e.g., texture changes, noise, inserting uninterested objects…). The specific alternation of non-shape features may differ by experiment
    \item $D_{aff}=[X_{aff},y_{aff}]$ by rotating or flipping $\left[X,y\right]$ so that the output image contains objects with different shapes from before (represent affine transformation of shape). 
    \item $D_{shuf}=[X_{shuf},y_{shuf}]$ by dividing $[X,y]$ into 16 $(4\times4)$ equally sized patches and shuffle their positions; (represent non-affine transformation of shape).
\end{itemize}

As shown in Fig. \ref{prob_img}, with these four datasets, we can compute a model’s SBI following the equations below. Let $IOU(D_{foo})$ denote the segmentation performance of the analyzed model on some dataset $D_{foo}$ ($D_{foo}$ is a placeholder). The relative performance drop caused by the digression of non-shape discriminative features in $D_{foo}$ from those in $D_{foo}$ can be quantified as:
$$PD_{foo}=1-\dfrac{IOU(D_{foo})}{IOU(D_{val)}}$$
We may obtain $PD_{rm},PD_{aff}$ and $PD_{shuf}$ in this manners by plugging in the place of $D_{foo}$ with $D_{rm},D_{aff}$ and $D_{shuf}$. Then, we smoothen their values through a Soft-Max function:
$$[\Delta_{rm},\Delta_{aff},\Delta_{shuf}]=SoftMax[PD_{rm},PD_{aff},PD_{shuf}]$$
Our SBI (Shape Bias Index) for the model to be assessed can now be computed as 
\begin{equation}
    SBI=\dfrac{\Delta_{aff}+\Delta_{shuf}}{\Delta_{rm}}
\end{equation}

If the performance drop caused by the changes to the shape  ($\Delta_{aff}$ and $\Delta_{shuf}$) is higher than that caused by changes in non-shape features ($\Delta_{rm}$), it reflects that the network is more sensitive to changes in shape than non-shape features. The value of SBI would then be larger than one and increases with higher sensitivity to shape changes and lower sensitivity to non-shape changes. The larger the SBI is, the more weight to shape information a network weighs in its inference process. 

In terms of computational complexity, during inference time, computing SBI takes one pass for each of the three probing sets $D_{rm}, D_{aff}$ and $D_{shuf}$, a relatively stable three-fold increase to inferences on $D_{val}$. In contrast, when generating the three probing sets from $D_{val}$, there would be significant variabilities. It could be as little as zero computing overhead if $D_{rm}$ was naturally occurring, to as much as iterating through VGG19 hundreds of iterations per image if we create $D_{rm}$ using algorithms like Neural Style Transfer~\cite{gatys2017controlling}. In most cases, however, the computational cost should be similar to performing data augmentation.

Since the framework for computing SBI requires a task-specific validation set and has broad room for alternative implementation of probing sets, it is an unnormalized metric. Direct comparisons between SBIs for different models require the same validation set and probing sets. An explicitly defined SBI threshold may be helpful for a qualitative answer to “whether a specific network is shape-learning.”

\section{Data}
\label{data}
To explore and verify factors that affect the shape learning of segmentation networks, we conducted an extensive set of experiments on a combination of synthetic and real images. We first use synthetic data to explore how the presence of different discriminative features, the size of the model receptive field, and the size of target objects influence shape learning. We then demonstrate the response of shape-learning segmentation networks to an increasing intensity of domain shift (e.g., brightness changes) and shape variations in target objects. 
In our synthetic dataset, each image contains one or two polygons sized 100-150 pixels in width and height as potential (target or non-target) objects. The objects can be simple or complex. Object(s) in each image is/are placed on some background with one random texture. The specific textures in each image, both for those associated with objects and backgrounds, are drawn from:
\begin{enumerate}
    \item the Colored Brodatz database (112 textures) for the training set, validation set $D_{val}$, and two of the three probing sets $D_{aff}$ and $D_{shuf}$.
    \item the CUReT database (61 textures) for images in probing set $D_{rm}$ so that no seen textures would exist in ($D_{rm}$)
\end{enumerate}
After the experiments and analysis of our synthetic datasets, we further validate our findings on three of real datasets. The three datasets are:
\begin{enumerate}
    \item BFGT: Breast and FGT MRI Dataset \cite{buda2020detection}
    \item FISH: A Large-Scale Dataset for Fish Segmentation and Classification\cite{ulucan2020large}
    \item LUNG: OSIC Pulmonary Fibrosis Progression\cite{OSIC_PFP,OSIC_PFP_msk} + COVID-19 CT scans\cite{glick,pavia, ma_jun_2020_3757476} 
\end{enumerate}

BFGT uses a subset of "Breast and FGT MRI Dataset"\cite{buda2020detection} containing slices parallel to axial planes of breast tomosynthesis. The masks for breast regions across different images are of similar shapes. 
FISH uses a subset of "A Large-Scale Dataset for Fish Segmentation and Classification" \cite{ulucan2020large} containing four species of fish in two genera with their photos pasted on a wood or marble surface. Each fish is positioned with its longest dimension positioned horizontally, second longest dimension positioned vertically, and shortest dimension positioned perpendicular to the surface.
Images and masks for LUNG are extracted from a combination of two different repositories. Specifically, "OSIC Pulmonary Fibrosis Progression" \cite{OSIC_PFP,OSIC_PFP_msk} and  "COVID-19 CT scans"\cite{glick,pavia, ma_jun_2020_3757476}. Both repositories contain images extracted from the axial view of lung X-ray images. The shapes of lungs across different images resemble each other. 

These three real datasets are purposely chosen following certain criteria to reproduce the configuration for computing SBI. Each dataset is selected to be containing objects of interest with similar shapes. However, within each dataset, there need to be at least two types of naturally occurring visual variability for those objects. For LUNG, the source of variability comes from the difference in the patient's health condition; for BFGT, the variability comes from the hardware vendor and filming parameters of the MRI machine; and for FSIH, the difference in species for those in the same family.

We divided each of the real datasets into four partitions: one as the training set, one as the  validation set, one noisy dataset created by adding six different types of noise to validation sets independently, and one with naturally occurring OOD data (e.g., images of fish from a different species but from similar family). These three versions of the datasets will help approximate the IOU($D_{rm}$) used for computing the SBIs for the respective models. 

\section{Implementation Details}
We conducted all our experiments with a UNet-like segmentation network. Unless otherwise specified, the implementation of our network is adapted from\cite{buda2019association}, with all batch normalization layers set to use per-batch statistics. We used Adam optimizer with lr=1e-3. Binary Cross Entropy loss was used as the loss function to train on both synthetic and real datasets. For the experiment elaborated in Section \ref{rf}, We replace all convolution layers in the encoding block of the model with atrous convolution layers that have dilation set to 2.

\section{Experiments}
\label{exp}
\begin{table*}[t!]
\footnotesize
    \centering
    \begin{tabular}{|c|c|}
    \hline
     \multicolumn{2}{|c|}{Annotation used for Synthetic Data (superscript 'c' means objects are complex)}\\
     \hline
     IOU(.)& the operation that computes the IOU of the model on the specified dataset\\
    \hline
    $D_{val}, D^c_{val}$& The dataset containing all discriminative features designed in the experiment group\\
    \hline
    $D_{rm},D^c_{rm}$ & The dataset with all possible non-shape features removed\\
    \hline
    $D_{aff},D^c_{aff}$ & The dataset obtained by applying affine transformation. Here we rotated $D_{val},D^c_{val}$\\
    \hline
    $D_{shuf},D^c_{shuf}$ & The dataset obtained by scramble and shuffle $D_{val},D^c_{val}$\\
    \hline
    $F_i, i\in\mathbb{N}$& Some unspecified discriminative features in our experiment design (e.g., shape, texture)\\
    \hline
    \multicolumn{2}{|c|}{Annotation used for Real Data}\\
    \hline
    $IOU_{val}$& performance of the model on the validation set\\
    \hline
    $IOU_{noisy}$& average performance of the model on the validation set with different types of noises added\\
    \hline
    $IOU_{OOD}$& performance of the model on data  containing naturally occurring visual variability.\\
    \hline
    \end{tabular}
    \caption{\footnotesize Symbols frequently used in this paper and a brief summary of their meaning}
    
\end{table*}

With a combination of synthetic and real data, we performed a comprehensive set of experiments to answer the following questions in the setting of semantic segmentation:
\begin{enumerate}
    \item Under what circumstances does a network learn shape?
    \item Do object size and model receptive field size affect shape learning?
    \item How do input perturbations affect the performance of shape-learning networks?
    \item Can augmentations encourage shape-learning with existing data?
\end{enumerate}
Question 1 is the principal question that we thoroughly examine in this paper. We answer this question using synthetic data representing simple objects (section \ref{simp_obj}) and complex objects (section \ref{cplx_obj}). Questions 2 and 3 are answered using synthetic data in Sections \ref{rf} and \ref{pert}, respectively. Finally, question 4 investigates how our systematic findings can be applied to encourage shape-learning and improve generalization in real data (Section \ref{real_data}).

Our experiment design follows a progressive manner. We start by listing a few non-overlapping concepts/features, including shape, that may potentially be useful for identifying objects of interest. Then, these listed features are permuted to form a power set of different combinations of features. We then compare and analyze how SBI changes as different features are included/excluded from the train data.

\subsection{Under what circumstances does a network learn shape? (simple object)}
\label{simp_obj}
\begin{figure}[ht]
\centering
\includegraphics[width =0.46\linewidth]{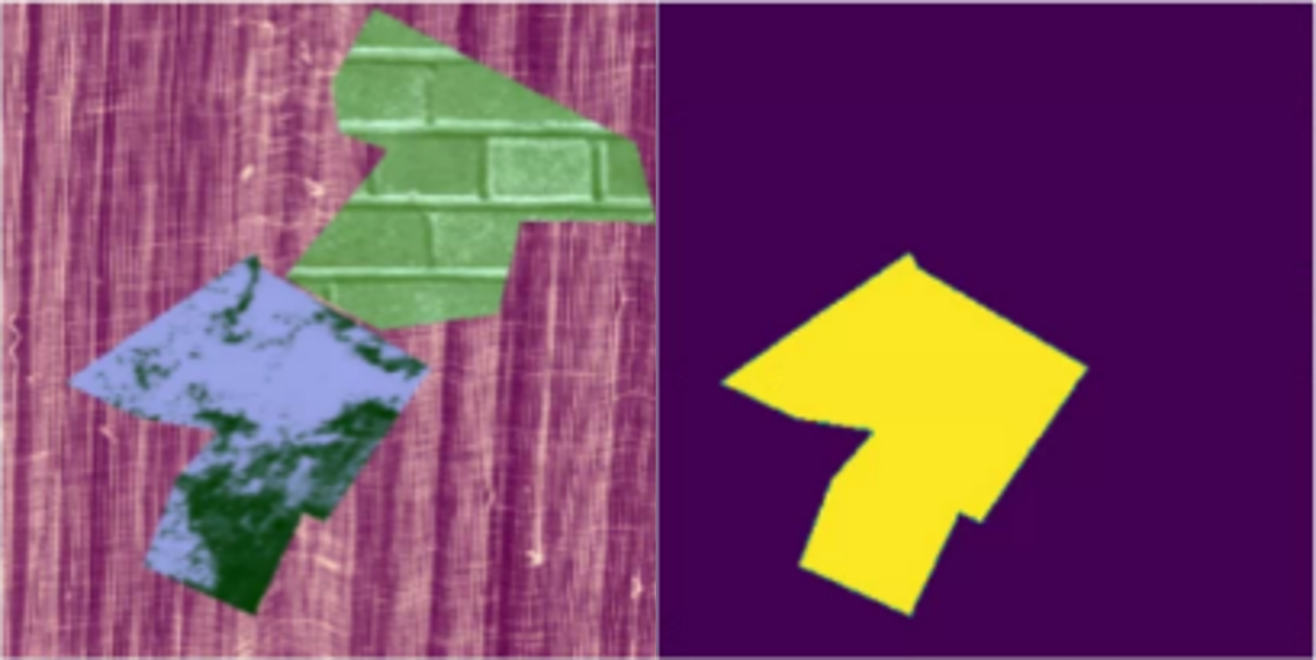}
\caption{\footnotesize Image of simple objects with only shape being discriminative}
\vspace{-2mm}
\label{shape_simp}
\end{figure}

We first answer the question on the capability of networks to learn simple objects’ shapes. In order to do so, we consider four different scenarios and four corresponding simulated datasets. In these scenarios, different features will be available based on which segmentation of the target object can be conducted. The first scenario is a simple one, where each image contains two simple objects in the shape of polygons: one with a discriminative shape (i.e., a shape that can be used to determine that this is the object of interest) throughout the dataset (target object) and the other with random shape (non-target object). An example training image is shown in Figure \ref{shape_simp}. The presence of other discriminative features such as texture in this initial training set is deliberately avoided. To ensure the model cannot use textures to classify, each object’s texture is randomly selected. 
In real-life scenarios, however, multiple discriminative features with similar predictiveness are usually present simultaneously (e.g., the shape of elephants and the color/texture of elephants’ skin). To explore how shape-learning behaviors are affected by discriminative features co-occurring with shape, we design the following additional discriminative features through which target objects in an image can be identified:	
\begin{enumerate}
    \item Texture: The texture of each pixel associated with target objects is drawn from a set containing five selected textures. In other words, a pixel is associated with the target object if and only if it has one of the selected textures. Non-target objects and the background have disjoint sets of textures from target objects.
    \item Singular: Instead of two polygons in each image, there is only one polygon of size 100-150 pixels in both dimensions, which is always the target object. Textures are randomly selected for both target and background unless co-occurred with texture. 
\end{enumerate}

\begin{figure}[ht]
\centering
\begin{subfigure}[b]{0.44\textwidth}
 \centering
    \includegraphics[width=\textwidth]{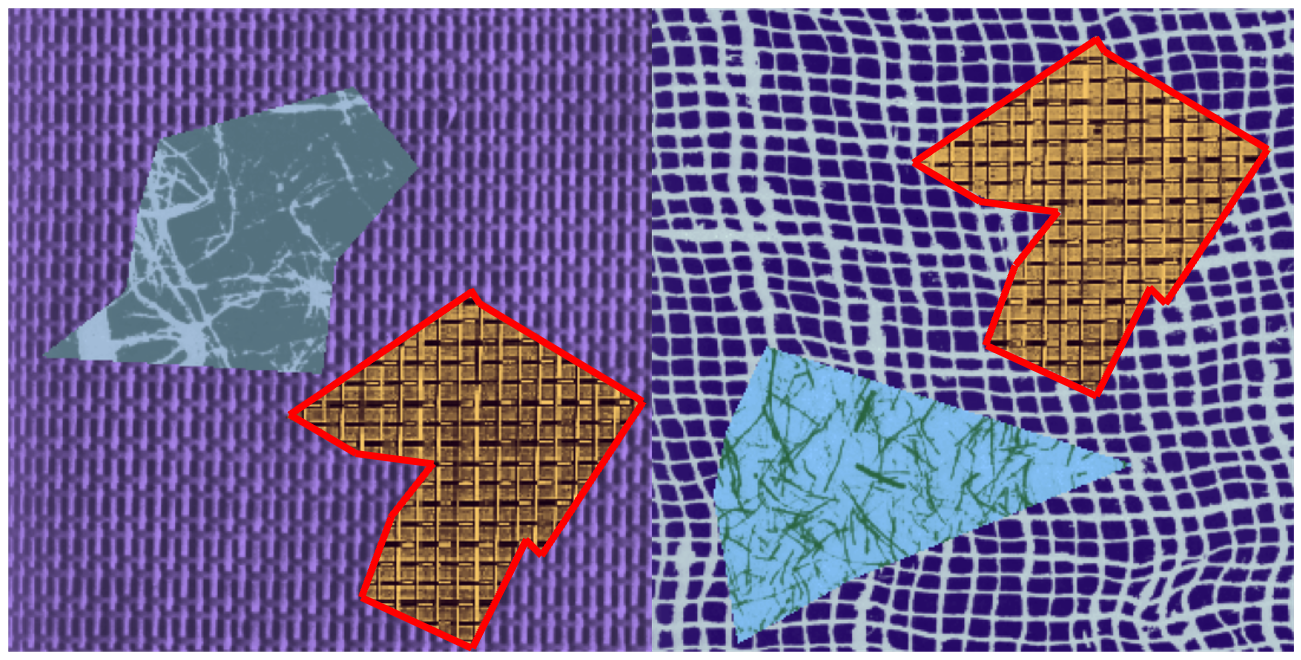}
    \caption{\scriptsize Texture: targets have specific textures}
\end{subfigure}
\begin{subfigure}[b]{0.44\textwidth}
 \centering
    \includegraphics[width=\textwidth]{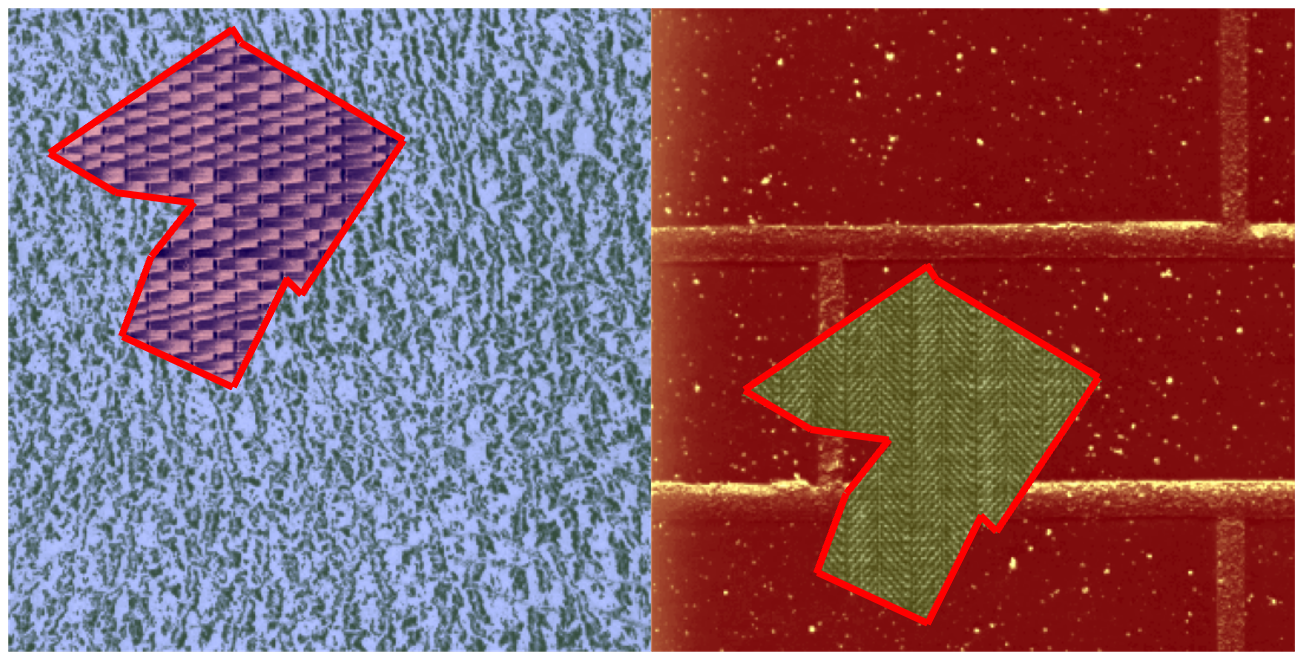}
    \caption{\scriptsize Singular: targets are stand-alone}
\end{subfigure}
\caption{\footnotesize Image of simple obj. with non-shape features (bounded in red)}
\label{non-shape_simp}
\end{figure}
Based on this definition of features, the remaining three scenarios, and corresponding simulated datasets are as follows. The second scenario is the same as the first one with the exception that texture is also a discriminative feature as described above. The third scenario is the same as the first with the addition of the “singular” discriminative feature as described above. In the fourth scenario, both texture and singular features are discriminative. We train a segmentation network on each training set. The performance of those models on $D_{val},D_{rm},D_{aff}\ and\ D_{shuf}$ (the probing datasets used to compute SBI), together with SBIs, is reported in Table \ref{simple_obj_table}.
\begin{table*}[t!]
\footnotesize
    \centering
    \begin{tabular}{|c|c|c|c|c|c|}
    \hline
    Discriminative features&IOU($D_{val}$) & IOU($D_{rm}$)&IOU($D_{aff}$)&IOU($D_{shuf}$)&SBI  \\
    \hline
    Shape only (simp. obj.)& 0.993&0.967&0.033&0.146&\textbf{4.490}\\
    \hline
    shape+texture&0.990&0.288&0.900&0.641&1.238\\
    \hline
    shape+singular&0.993&0.589&0.897&0.434&1.902\\
    \hline
    shape+texture+singular&0.999&0.398&0.976&0.812&1.222\\
    \hline
    \end{tabular}
    \caption{\footnotesize Performance of UNets trained on datasets containing simple objects with different discriminative features on the probing sets and their SBIs. The raw dataset and $D_{val}$ are assumed to contain shape+texture+singular as discriminative features.}
    \vspace{-5mm}
    \label{simple_obj_table}
\end{table*}

From the last column of Table \ref{simple_obj_table}, one can clearly see that only the model from the first scenario, where the shape is the only discriminative feature of the target object, was able to show considerable shape learning behavior (high SBI). Whenever another feature was available to aid the segmentation, the models did not show the tendency to learn shape (low SBI).

Analysis of performance on the specific datasets (columns 2-5) allows for more detailed insights. First, the model trained on the dataset where shape was the only discriminative feature achieved high segmentation performance on the validation set $D_{val}$. It demonstrates that a segmentation network  can recognize shapes because objects’ shapes are the only feature allowed for identifying the object of interest. Recall that $D_{rm}$ perturbs non-shape features while $D_{aff}$ and $D_{shuf}$ perturbs target objects’ shapes. This model experiences a lower performance drop when tested on $D_{rm}$ than on $D_{aff}$ and $D_{shuf}$. The model’s robustness against significant texture changes (also brightness changes and noise) and brittleness against shape changes aligns with our intuition about the properties of a shape-learning network. We hence conclude segmentation networks are capable of learning discriminative shape information in the target objects for simple objects. 

We then observe the behavior of models trained on datasets containing other discriminative features in addition to shape (i.e., shape+texture, shape+singular, and shape+texture+singular). The discriminative features in addition to shape unanimously lower the SBI of each model to a significant extent, despite their similar validation IOUs. This phenomenon indicates that shape-learning is not a default behavior of segmentation networks; rather, these findings suggest that they default to learning other discriminative features if present. Practically, the suppression or removal of features unrelated to shape may be vital to encouraging the shape learning of networks in a typical object segmentation task. 

It is worth noting that we observed that the model in scenario 3 (shape+singular) generally segmented both objects in the image. We believe this is because shape+singular was trained on a single foreground object, which may have encouraged the model to identify “any foreground objects of similar size as the targets seen in the training set”, even if they have completely different shapes.

Based on the results from the experiments in this section, we draw the following conclusions:
\begin{enumerate}
    \item A UNet-like network does not prioritize learning shape (by default) in the presence of other discriminative features of similar predictive capability (e.g., a unique texture (texture), or being the only object in the image (singular)). 
    \item A segmentation network learns shape only if the target objects’ shapes are the sole discriminative feature.
\end{enumerate}
In the next section, we will show that these conclusions extend to complex objects. We will also explore the interactions between shape and other discriminative features. 
\subsection{Under what circumstances does a network learn shape? (complex object)}
\label{cplx_obj}
In the previous section, we investigated shape learning in simple objects. We now extend our experiments to the case where targets are complex objects, a setting more closely resembling real-life scenarios. Like in the previous section, we first explore whether there exists a scenario in which a network exhibits a shape-learning behavior for complex objects. Mimicking the previous experiment, we train a segmentation network on a dataset containing only shape as the discriminative feature. 
\begin{figure}[ht]
\centering
\includegraphics[width =0.46\linewidth]{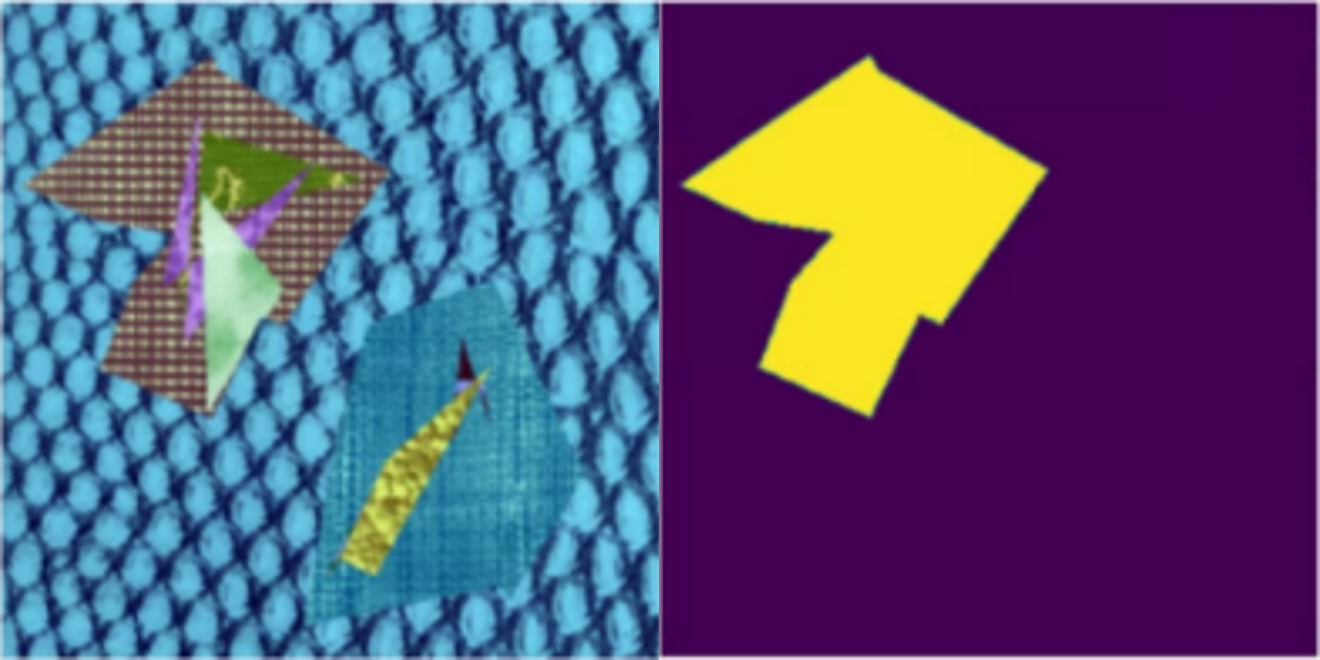}
\caption{\footnotesize Image of complex objects with only shape being discriminative}
\vspace{-1mm}
\label{shape_cplx}
\end{figure}

We then generate different combinations of discriminative features and add them to the dataset and observe how they affect SN’s ability to learn and leverage shape information. Since we assume targets are complex objects, each image can now contain more sophisticated features. In addition to “singular” and “texture” modifications in the simple objects described in Section \ref{simp_obj}, we also introduce “structure” and “semi-singular” as potential discriminative features:
\begin{enumerate}
    \item Structure: All simple objects within the target have the same shape and are arranged in a fixed relative position to each other. Their respective textures, unless otherwise specified, are randomly sampled. 
    \item Semi-singular: There are two polygons sized 100-150 pixels in width and height per image. The target object is the only complex object in each image. The structure and texture of target objects are random. Note that semi-singular and singular are two features that cannot appear simultaneously, as semi-singular assumes the presence of at least two objects of similar sizes to the target object in the image. 
\end{enumerate}
\begin{figure}[ht]
\centering
\begin{subfigure}[b]{0.44\textwidth}
 \centering
    \includegraphics[width=\textwidth]{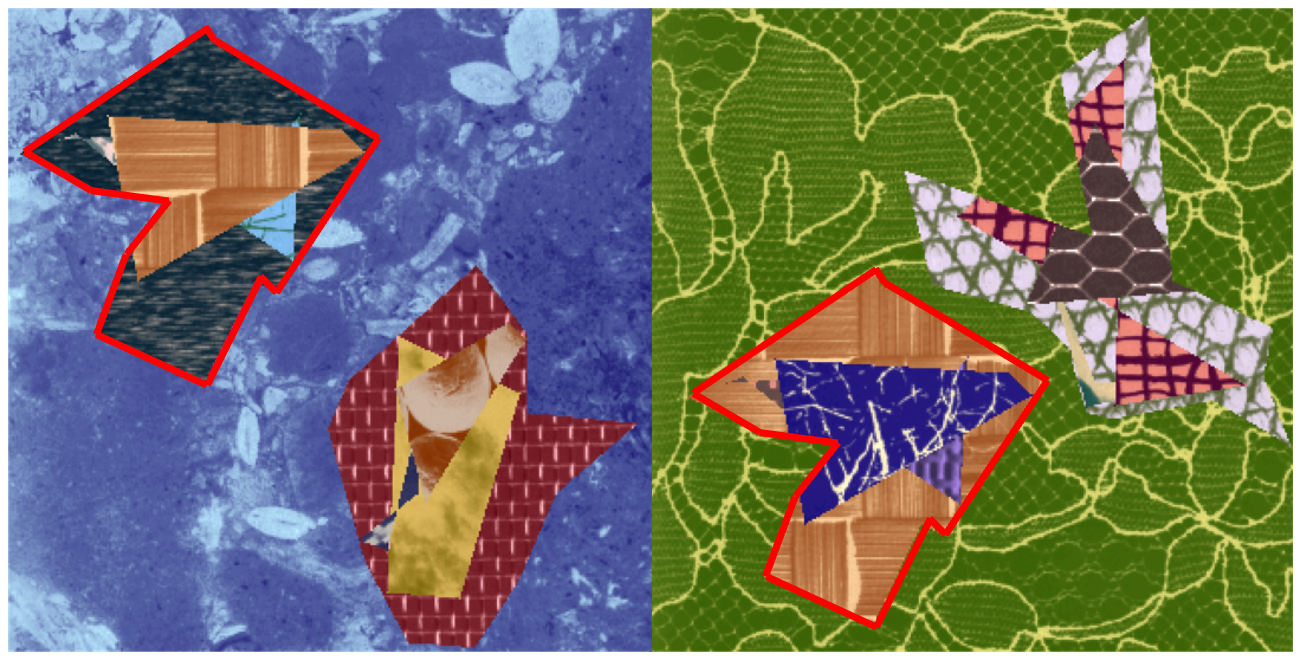}
    \caption{\scriptsize Structure: targets have the same layout of sub-components}
\end{subfigure}
\begin{subfigure}[b]{0.44\textwidth}
 \centering
    \includegraphics[width=\textwidth]{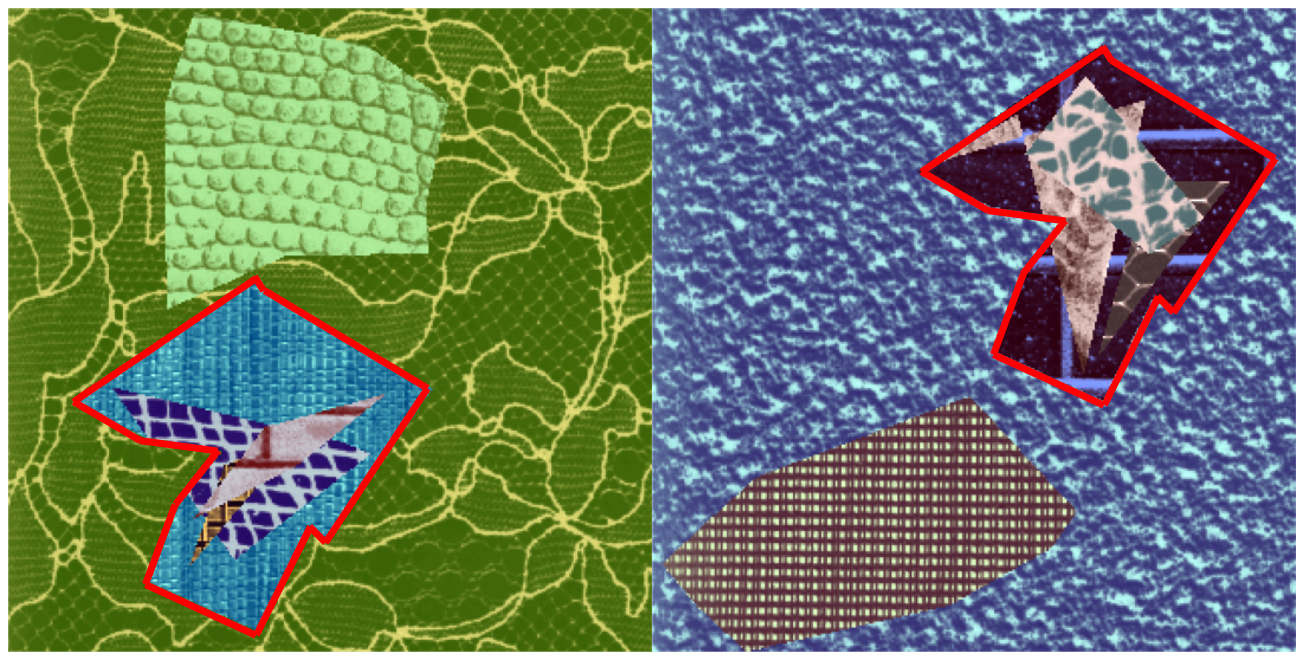}
    \caption{\scriptsize semi-singular: targets are complex obj., while non-targets are simple obj.}
\end{subfigure}
\caption{\footnotesize Image of complex objects with non-shape features (bounded in red)}
\vspace{-1mm}
\label{non-shape_cplx}
\end{figure}

\begin{figure*}[t]
\centering
\includegraphics[width =\linewidth]{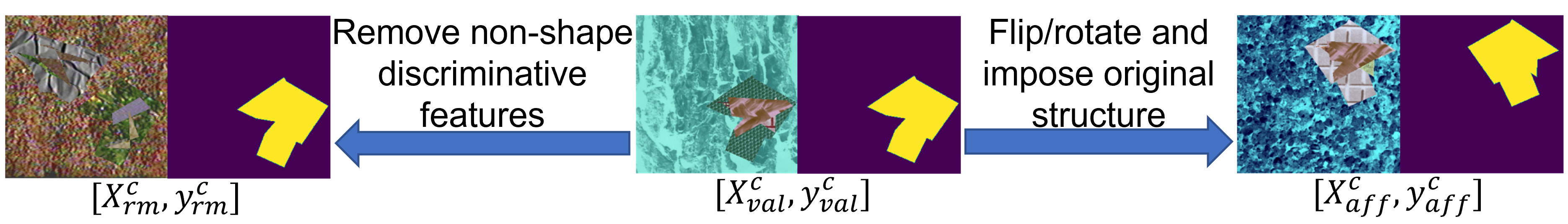}
\caption{\footnotesize Image of complex objects with only shape being discriminative}
\vspace{-4mm}
\label{prob_cplx}
\end{figure*}
As the combination of discriminative features becomes more complicated in this experiment, we adjusted the data generation process for $D_{val}, D_{rm},D_{aff}$ and $D_{shuf}$ to maintain the efficacy of SBI. To avoid confusion, datasets in this experiment to compute SBI are denoted as $D_{val}^c,D_{rm}^c,D_{aff}^c$ and $D_{shuf}^c$, distinguished from those in the prior section by superscript ‘c’ (stands for “complex"). $D_{val}^c$ is generated to simultaneously contain all listed discriminative features in each image, namely singular, texture, and structure. $D_{rm}^c$ is generated to contain both target and non-target complex objects with random structures in textures randomly drawn from CUReT database; shape is the only discriminative feature through which target objects may be identified. $D_{aff}^c$ is obtained through rotating/flipping the target objects in $D_{val}^c$ while keeping the structure, texture, and approximate size of the rotated object unaltered. A sample of the three datasets each is listed in Fig. \ref{prob_cplx}. $D_{shuf}^c$ divides $D_{val}^c$ into patches and shuffled them. 

Table \ref{complex_obj_table} reports the performance of each network trained on the baseline with only shape being discriminative, and on datasets with different combinations of additional discriminative features.
\begin{table*}[t!]
    \centering
    \footnotesize
    \begin{tabular}{|c|c|c|c|c|c|}
    \hline
    Discriminative features&IOU($D_{val}^c$) & IOU($D_{rm}^c$)&IOU($D_{aff}^c$)&IOU($D_{shuf}^c$)&SBI  \\
    \hline
    Shape only (complex obj.)& 0.996&0.933&0.266&0.278&\textbf{3.940}\\
    \hline
    shape+singular&0.987&0.522&0.857&0.614&1.628\\
    \hline
    shape+semi-singular&0.983&0.637&0.758&0.572&1.952\\
    \hline
    shape+texture&0.985&0.254&0.958&0.756&1.098\\
    \hline
    shape+structure&0.998&0.474&0.589&0.466&1.896\\
    \hline
    shape+singular+texture&0.979&0.456&0.948&0.740&1.358\\
    \hline
    shape+singular+structure&0.999&0.463&0.825&0.6456&1.526\\
    \hline
    shape+texture+structure&0.999&0.270&0.905&0.690&1.180\\
    \hline
shape+texture+singular+structure&1.000&0.458&0.928&0.761&1.356\\
    \hline
    \end{tabular}
    \caption{\footnotesize Performance of UNets trained on datasets containing complex objects with different discriminative features on the probing sets and their SBIs}
    \vspace{-4mm}
    \label{complex_obj_table}
\end{table*}

According to Table \ref{complex_obj_table}, the networks continue to show a good capability to utilize discriminative shape information for segmentation tasks, as indicated by the high SBI and $D_{val}^c$. It also has a significantly smaller performance drop on $D_{rm}^c$ than on $D_{aff}^c$ or $D_{shuf}^c$. We thus conclude that segmentation networks are capable of segmenting through shape, regardless of the targets being simple or complex objects.

Similar to in experiments with simple objects, we again see that the features “texture” and “singular” in the training set distract a network from recognizing complex objects by shape. “semi-singular” appears to have similar effects as “singular” based on its performance on each sub-task. As we consider shape+singular teaches a network to recognize any object surrounded by pixels with only one texture representing the background, shape+semi-singular may result in a network identifying any complex object surrounded by pixels with only one texture representing the background. 
\begin{figure}[ht]
    \centering
    \includegraphics[width=0.5\linewidth]{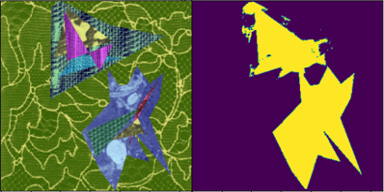}
    \caption{\footnotesize Network trained on shape+singular shows a significant failure to exclude objects with undesired shape.}
    \label{semi-singular-fp}
\end{figure}

The image on the left of Fig. \ref{semi-singular-fp} contains two polygons: one with the desired shape and another with an irrelevant shape. The two polygons form two complex objects in the background, with only the one in the lower-right corner being the target. However, the network trained on “+semi-singular” tends to also consider irrelevant objects as targets.
As we include more types of discriminative features in this group of experiments, we also found the following patterns. Let $\{Shape,F_0,\ldots,F_i\}$ be all possible discriminative features presenting in the training set, where $\{F_0,\ldots,F_i\}$ are some non-shape features. 
In most cases, including $\{Shape,\ F_0,\ldots,F_j\}\cup\{Shape, F_{j+1},\ldots,F_i\}$ in  the training of network (e.g., shape+singular+texture) leads to a network with SBI in-between those of networks trained on $\{Shape, F_0,\ldots,F_j\}$ and $\{Shape, F_{j+1},\ldots,F_i\}$ respectively (e.g., shape+singular and shape+texture). 
In less common cases, such as shape+singular+structure, the multiple non-shape features may collaboratively cause more severe distraction of shape-learning either of it can do alone. However, in no circumstance does a network trained on $\{Shape,F_0,\ldots,F_i\}$ have higher SBI than on both $\{Shape,\ F_0,\ldots,F_j\}$ and $\{Shape,F_{j+1},\ldots,F_i\}$. This observation reflects that though shape is generally ignored in the presence of other discriminative features, other non-shape features may be learned collaboratively, rather than one preferred over another. This finding amends the hypothesis made in the study of Hermann et al. \cite{hermann2020shapes} that neural networks primarily utilize only one discriminative feature at a time, but only in rare cases multiple features simultaneously. Instead, our findings suggest the simultaneous learning of multiple features is prevalent.  

Based on the results from experiments on complex objects, we extend our arguments made in Section \ref{simp_obj} to both the case of simple and complex objects. We also draw additional conclusions:
\begin{enumerate}
    \item A segmentation network does not prioritize learning shape in the presence of other discriminative features with similar prediction capabilities. This statement holds for both simple and complex objects.
    \item Though shape learning is distracted in the presence of other discriminative features, other discriminative features tend to be learned simultaneously if multiple of them are present.
\end{enumerate}

Recall that the discriminative features were progressively added to our experiment. For each name discriminative feature, the effects of its presence or absence are well-recorded. Such a setting effectively constitutes an ablation study, allowing us to compare and infer hypotheses on CNN's shape learning behavior.

\subsection{How do object size and model receptive field affect shape learning?}
\label{rf}
In the previous sections, we learned that networks are capable of learning shape, though doing so requires the absence of undesired discriminative features. We now explore how the size of target objects and model receptive fields may affect shape learning. Geirhos et al.\cite{geirhos2018imagenet} reported that restricting the receptive field size of ResNet would prevent it from converging on stylized ImageNet. Model variants with larger receptive fields seem to suffer less from this phenomenon. This observation suggests a potential correlation between receptive field size and the tendency or capability of CNNs to leverage shape information. 

We further expand on this topic by systematically analyzing how the size of an object of interest and the model receptive field affect shape learning. By receptive field, we mean the maximum possible number of pixels that can be involved in the computing of a pixel at the bottleneck of the network. 

We resized the training set for model shape only (simple obj.) mentioned in Fig. \ref{shape_simp} to 11 different sizes ranging from $160\times160$ to $480\times480$ with an interval of 32 pixels. We pair these 11 differently sized training sets each with two UNets of slightly different model architecture. Specifically, the CNN layers in the encoder of one UNet (denoted as UNet\_140) all have dilation=1, while the other UNet (denoted as UNet\_210) set the CNN layers in encoder block 2,3,4 to have dilations=2. This allows an expansion of receptive field size without changing the total number of parameters for the models. The formula for computing the receptive field size is:
\begin{itemize}
    \item UNet\_140: ((((1+4)*2+4)*2+4)*2+4)*2+4 = 140
    \item UNet\_210: ((((1+4)*2+9)*2+9)*2+9)*2+4 = 210
\end{itemize}

\begin{figure}[ht]
    \centering
    \includegraphics[width=0.65\linewidth]{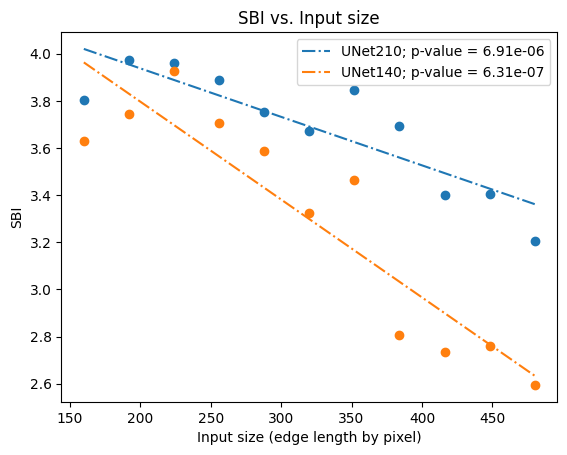}
    \caption{\footnotesize SBIs for UNets with different receptive field sizes when trained on images containing target objects of different sizes. The p-value listed is the p-value for the f-test on the linear regression.}
    \vspace{-1mm}
    \label{rf_trend}
\end{figure}

We train each [model variant, input size] pair with SGD optimizer (lr = 1e-2, weight\_decay= 1e-5, momentum=0.9) until the models fully converge. As shown in Fig. \ref{rf_trend}, for all input sizes, UNet\_210 has a larger SBI than UNet\_140. Plus, a clear and sudden drop in SBI as the input size goes larger than 382 is observed for UNet\_140, and 420 for UNet\_210, when the object size inside each image becomes larger than the models' receptive field sizes. 

A statistically significant (p-value<0.05 for F-test) trend of decrease is observed for both UNet\_140 and UNet\_210 as the input image size increases. We hence conclude that models with small receptive fields may have difficulty learning shape. Increasing the model receptive field and/or reducing ROI size may alleviate this phenomenon. This increase in SBI would diminish once the receptive field size versus object size reaches a certain threshold. 
Despite increasing model receptive field size and/or reducing ROI size may encourage shape-learning, there are drawbacks to both operations. Increasing a model’s receptive field usually introduces more parameters to it. More parameters may lead to longer training time and a higher risk of overfitting. Using dilated convolutional layers is a potential solution for increasing the model receptive field size without changing parameter sizes. Reducing ROI size reduces image resolution, resulting in the loss of potentially useful shape information. In practice, the hyperparameter turning is an essential step for obtaining a shape-learning segmentation network with high segmentation performance (e.g., IOU, DSC) on its designated task.

\subsection{How do input perturbations affect a shape-learning segmentation network? }
\label{pert}
In the previous sections, we explored how the presence of non-shape discriminative features, varying receptive field size, and varying object size can affect a network’s shape learning. We now investigate how a shape-learning network responds to feature variations that may be common in a real-life scenario. We still use synthetic data to approach this question and control the degree of feature variation. Specifically, we observe how a shape-learning network responds to brightness changes and shape variations.
\subsubsection{Perturbations on brightness}
Brightness change is one of the most prevalent domain shifts in real life that may significantly alter the textures of objects in images. Zhe et al.~\cite{zhu2021deep} demonstrated that models trained on data with limited variability in object brightness have poor generalizability to data with other brightness levels. Following common intuition, if a network learns shape, it should be insensitive to the changes in textures, hence be robust to changes in brightness. 

\begin{figure}[h!]
\centering
\begin{subfigure}[b]{0.49\textwidth}
 \centering
    \includegraphics[width=\textwidth]{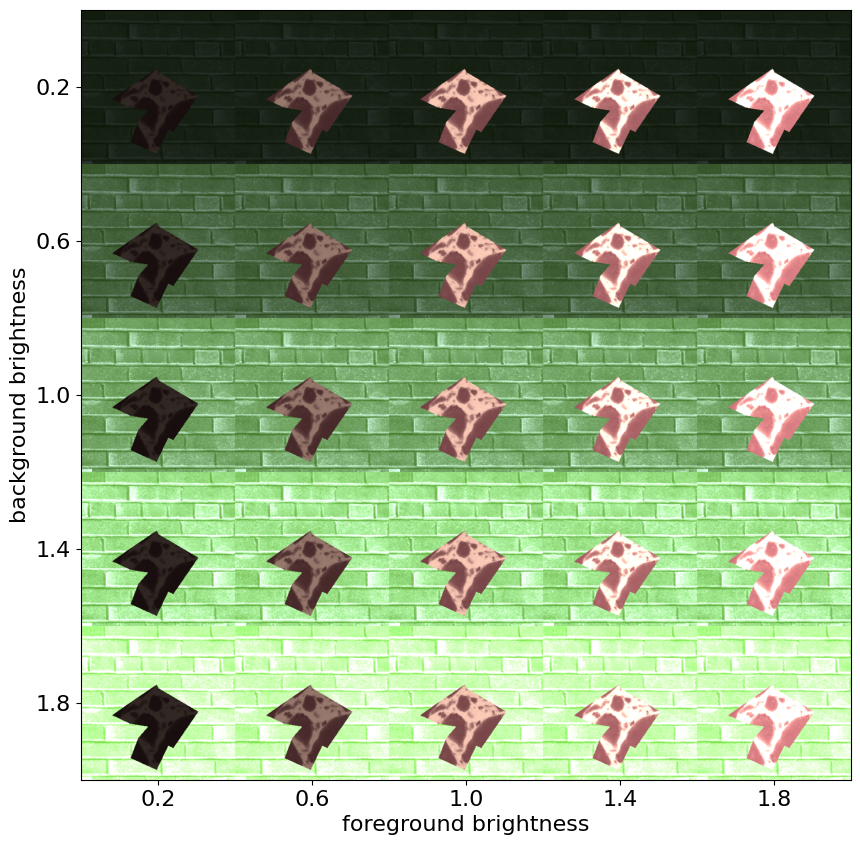}
\end{subfigure}
\begin{subfigure}[b]{0.49\textwidth}
 \centering
    \includegraphics[width=\textwidth]{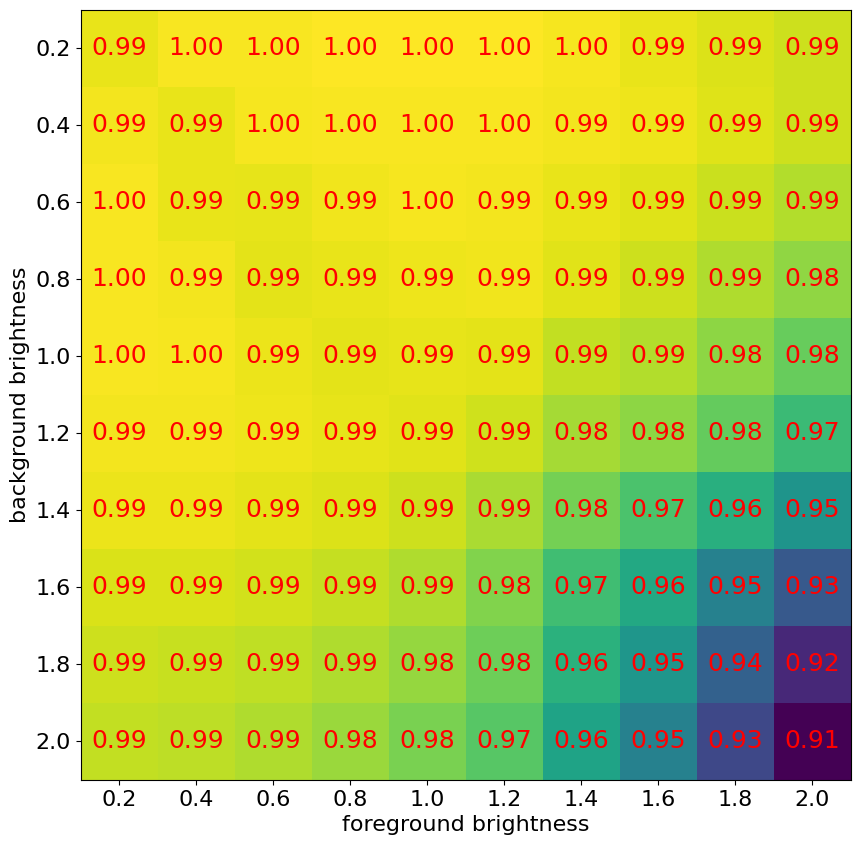}
\end{subfigure}
\caption{\footnotesize IOU of shape-learning UNet on dataset each with different combinations of foreground and background brightness change.}
\vspace{-5mm}
\label{brightness_pertub}
\end{figure}

We test this assumption by applying 100 unique combinations of foreground and background brightness pairs to the validation set and observe how our shape-considering model responds to the performance changes. Results in Fig. \ref{brightness_pertub} show that the drop in model performance, measured by IOU, falls within the 0.9-1 range. This range is significantly smaller than the range reported by Zhe et al. We hence conclude that a shape-learning network is robust against changes in brightness, as well as other forms of mild domain shifts.
\subsubsection{Perturbations on shape and textures}
In Section \ref{simp_obj}, we show a segmentation network capable of utilizing unvaried shape information in the absence of other discriminative features. However, in real-life scenarios, two different objects rarely have identical shapes even if they are in the same category. If a shape-learning network shows reasonable performance only when an identical shape appears, it has little advantage over symbolic methods such as template matching. We hence explore how a network trained on identical shapes responds to increasing degrees of shape variation. We also analyze the effects of including shape variation in the training set on SN’s shape learning behavior. 

\begin{figure*}[t!]
    \centering
    \includegraphics[width=\linewidth]{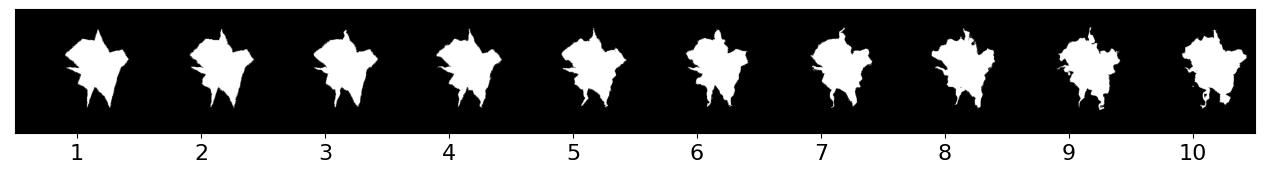}
    \caption{\footnotesize Ground-truth masks of target objects after applying different intensities of elastic deformation}
    \vspace{-1mm}
    \label{elastic_msk}
\end{figure*}

In doing so, we take $D_{val}$ used in for simple objects and apply increasing degrees of elastic deformation to the original polygon masks to create different levels of shape variations in the target objects, as shown in Fig. \ref{elastic_msk}. We then denote the dataset derived from $D_{val}$ as $\{D_{val}^1,\ldots, D_{val}^{10}\}$, where the superscript denotes the degree of shape variation. We test how models trained on shape only (simple obj.), presented in Section \ref{simp_obj} respond to such variations. Then, we train three additional models on three modified versions of shape only (simple obj.), each with different degrees of shape variation injected. After training, all four networks with different extents of shape variation (from no variation to some variation) show strictly decreasing IOUs as the degree of elastic deformation increases. 
\begin{figure}[ht]
    \centering
    \includegraphics[width=0.7\linewidth]{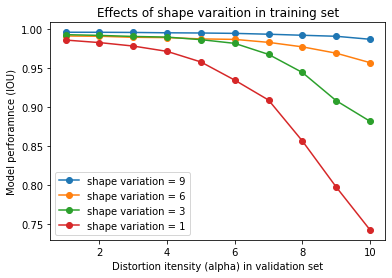}
    \caption{\footnotesize Performance of shape-learning UNets trained with training sets containing different degrees of shape variation on shape variations in validation sets}
    \vspace{-3mm}
    \label{elastic_result}
\end{figure}

According to Fig. \ref{elastic_result}, models trained with little shape variation have much steeper decreases in performance when tested on perturbed input. These results show that shape-learning networks develop tolerance to shape variations according to variations in target objects’ shape during training. On the other hand, the tolerance of shape-learning networks to shape variation may be affected by the extent of shape variation contained in the training set.
\subsection{Validation of finds on real data and exploration of augmentations}
\label{real_data}
In the previous subsections, we explored factors affecting the shape learning of networks with synthetic data. In this section, we demonstrate how those findings can be applied to real-life tasks. Specially, we assess how well our conclusions on synthetic data align with real-world datasets combined with data augmentation, a common technique used in machine learning to improve the generalizability of models on OOD data. 

Additionally, we evaluate augmentation methods to encourage shape learning when shape is a good discriminative feature in the dataset. Our selection of potential augmentation methods is inspired by the conclusions drawn from the systematic study with the synthetic data regarding the conditions under which segmentation networks learn shapes.
We conduct our experiments on three real datasets that contain targets of similar shape, as listed in Section \ref{data}. Each dataset is split into four components: 
\begin{enumerate}
    \item Training set
    \item Validation set (serve as $D_{val}$), which has a similar appearance to the training set. The performance of model on this dataset is denoted as $\mbox{IOU}_{val}$
    \item Noisy data (serves as part of $D_{rm}$) obtained by separately adding six different types of noise to $D_{val}$. The six types of noise are Gaussian noise, shot noise, impulse noise, defocus blur, pixelate, and motion blur. Note that though adding noise may also be a common technique used in data augmentation, we did not inject noise into training data to prevent data leaking. The performance of model on this dataset is denoted as $\mbox{IOU}_{noisy}$
    \item OOD data (serves as part of $D_{rm}$) with naturally occurring domain shifts, which contains targets of similar shape as those in the training set, but with different visual appearance. The performance of model on this dataset is denoted as $\mbox{IOU}_{OOD}$ 
\end{enumerate}

Note that the concept of OOD images can also be generalized to images with hyperspectral variabilities. Each channel of hyperspectral images contains intensity information of a specific spectral wavelength~\cite{857802}. Ideally, one can identify different materials (known as endmembers) based on their unique spectra signatures. However, in practice, mixtures of spectra of materials and the environmental conditions (e.g., local temperature and humidity and atmospheric effects) may shift the pixel intensity at each wavelength channel by the time of reaching hyperspectral camera~\cite{8528557}. The information changed in such a variation roughly follows our definition of \emph{texture} in \ref{def}. 

Recall that three probing sets in addition to the validation set are needed to computer SBI. The other probing datasets $D_{aff}$ and  $D_{shuf}$ are obtained following the same workflow as shown in Section \ref{def}. 

For each training set, we apply the same group of augmentations, which may be perceived as perturbations to certain discriminative features as below:
\begin{enumerate}
    \item Color Jitter
    \item Separate Color Jitter
    \item Neural Style Transfer (NST)
    \item Negative Insertion
    \item Random Resized Crop
    \item Random Crop Reflect
\end{enumerate}
Most of the augmentations are used in an attempt to suppress learning texture features. Specifically, Color Jitter changes the brightness, contrast, and saturation of images. Separate Color Jitter is an extension of Color Jitter which uses different parameters for targets and the rest of the image.

Neural Style Transfer (NST) \cite{gatys2017controlling} changes the Gram matrix of feature vectors. Visually, they significantly alter the texture of images but barely change objects' shape.

Negative Insert is designed to suppress singular; it inserts objects with similar textures but different shapes from the targets. In our case, this operation is implemented by cropping the targets into multiple patches and then shuffling and pasting them back to the original image. The inserted objects are not labeled as targets. 

Random Resized Crop may have complicated effects, as it suppresses both shape and texture – the cropping changes the targets’ shapes while the resizing changes texture density. Random Crop Reflect changes only object shape while preserving the texture on each object by padding the cropped patch reflectively. If target objects are mirrored, their masks are also mirrored. 

Following our conclusions in Section \ref{simp_obj} and Section \ref{cplx_obj}, we expect all augmentations to produce higher model SBI than the baseline (no augmentation), except for Random Resized Crop and Random Crop Reflect, as these two augmentations alter the shape of target objects. 
\subsubsection{Validation with BFGT}
The complete BFGT dataset contains 922 volumes of breast MRI images produced by GE and SIEMENS machines. We obtain 219 slices (up to 3 slices per volume) produced by the GE machine as our training and validation set. We then select 81 slices (up to 3 slices per volume) produced by SIEMENS as our out-of-domain dataset. Each image is resized to have the same dimension of $256\times256$. We report below the mean IOUs and SBI of models trained on data with different augmentations. 
\begin{table}[ht]
\footnotesize
    \centering
    \begin{tabular}{|l|l|l|l|l|}
    \hline
    Training set &	$\mbox{IOU}_{val}$&$\mbox{IOU}_{noisy}$&$\mbox{IOU}_{OOD}$&SBI\\
    \hline
    No Augmentation&0.853&0.718&0.761&3.396\\
    \hline
    Color Jitter&0.830&0.744&0.820&3.726\\
    \hline
    NST&0.850&\textbf{0.802}&\textbf{0.823}&\textbf{4.102}\\
    \hline
    Sep. Color Jitter&0.872&0.730&0.780&3.696\\
    \hline
    Negative Insertion&0.867&0.717&0.748&3.996\\
    \hline
    Random Resized Crop&0.858&0.623&0.725&2.538\\
    \hline
    Random Crop Reflect&0.863&0.619&0.774&2.518\\
    \hline
    \end{tabular}
    \caption{\footnotesize Performance of models trained on datasets with different augmentations when applied to the validation set, dataset with noise injected, and dataset with naturally occurred domain shifts for BFGT}.
    \vspace{-6mm}
    \label{BFGT_rslt}
\end{table}

In experiments with BFGT, augmentations with no modifications on targets’ shapes achieved higher SBIs than the baseline model, while those that altered targets’ shapes resulted in models with smaller SBIs. This phenomenon matched our prediction. Color Jitter (brightness shifts for grayscale images), Sep. Color Jitter, and NST all lead to models with higher performance on OOD and Noisy data. In contrast, Negative Insertion showed no improvement in performance on OOD nor on Noisy data compared to the baseline, despite its higher SBI. This is because SBI also takes an SN’s performance on objects with dissimilar shapes into consideration. Random Resized Crop and Random Crop Reflect alters shape information in the training set. Both models have smaller SBIs than the baseline, plus similar or lower performance on OOD and noisy data. The results confirm that augmentation not altering shape improves models’ SBIs; Color Jitter and NST are apparently the most practical augmentation methods to be applied to encourage shape learning and improve models’ generalizability. 

\begin{figure}[ht]
\centering
\begin{subfigure}[b]{0.24\textwidth}
 \centering
    \includegraphics[width=\textwidth]{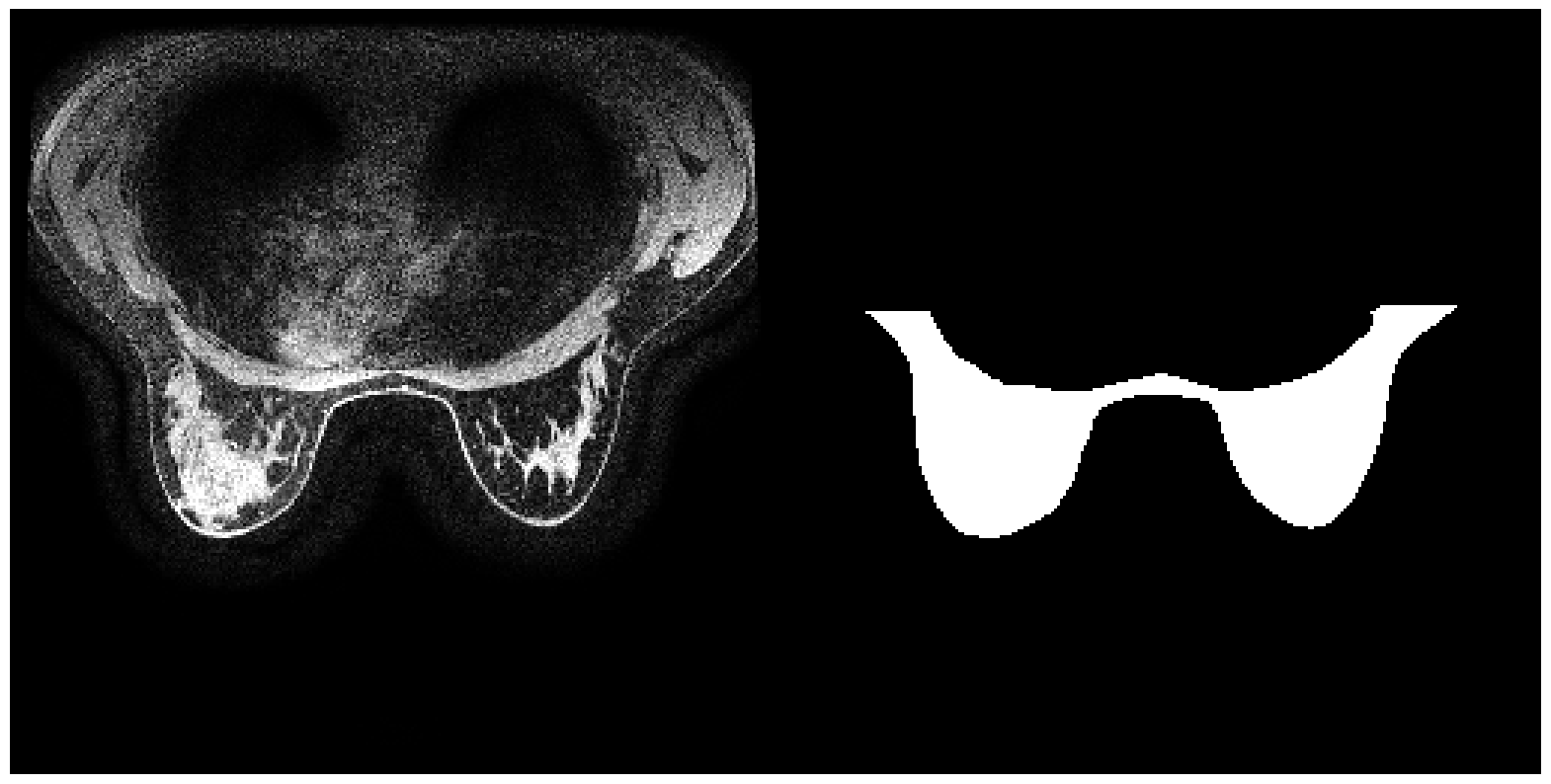}
    \caption{\scriptsize Image from GE machine}
\end{subfigure}
\begin{subfigure}[b]{0.24\textwidth}
 \centering
    \includegraphics[width=\textwidth]{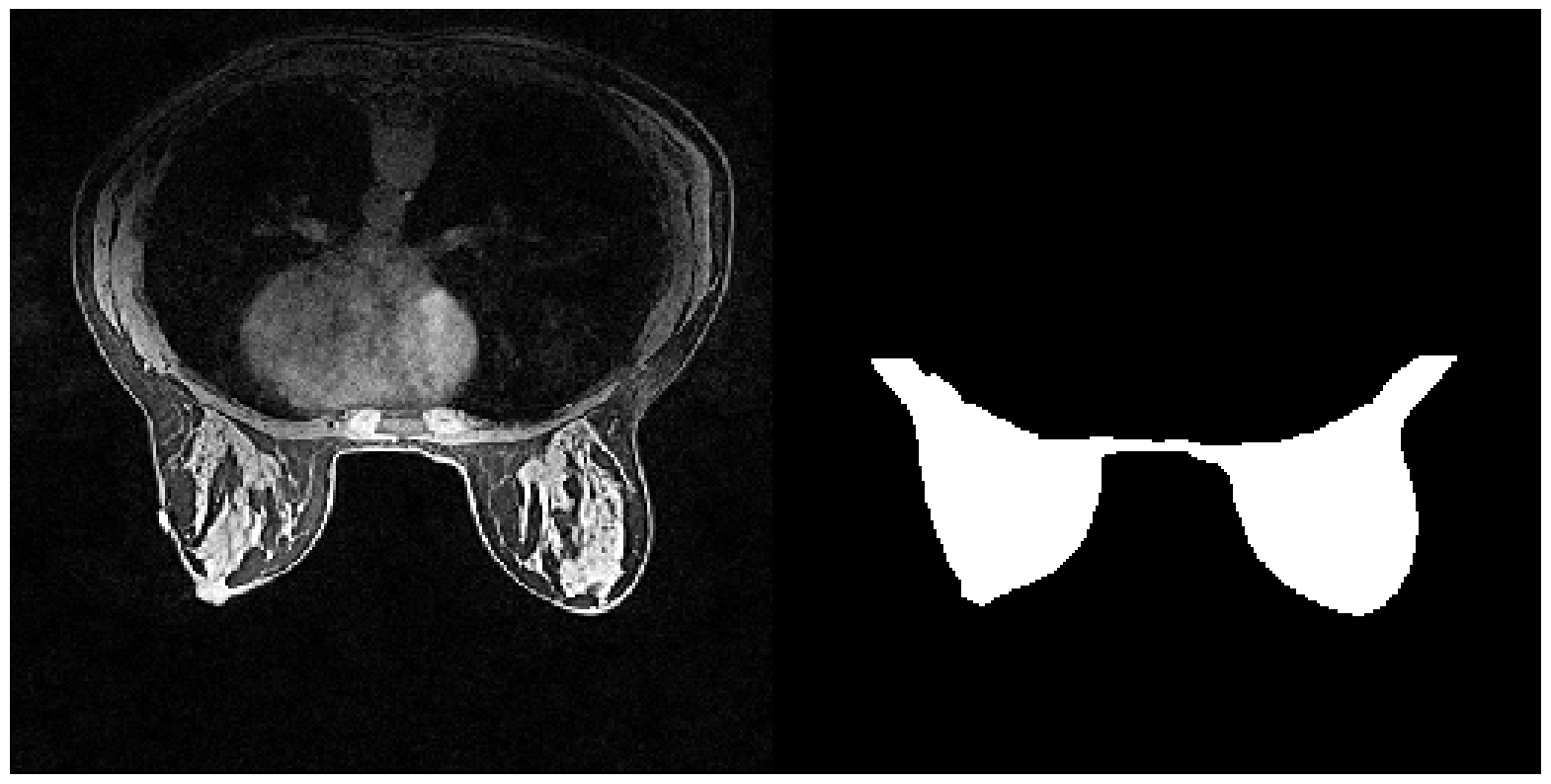}
    \caption{\scriptsize Image from SIEMENS machine}
\end{subfigure}

\begin{subfigure}[b]{0.48\textwidth}
 \centering
    \includegraphics[width=\textwidth]{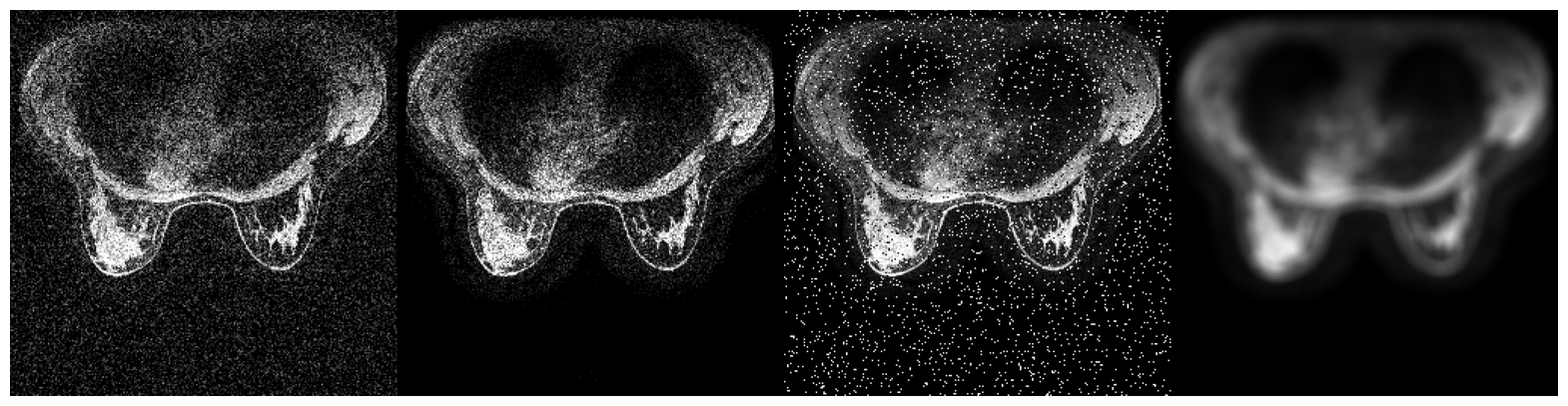}
    \caption{\scriptsize The image collected from GE machine with (from left to right) Gaussian noise, shot noise, impulse noise, and defocus blur applied.}
\end{subfigure}
\caption{\footnotesize Samples from the three partitions of data domain in BFGT}

\label{BFGT_samples}
\end{figure}

\subsubsection{Validation with FISH}
The fish dataset contains fish of various species photographed from similar perspectives. We extract fish of 4 species in two families, where species in the same family tend to have similar shapes. Two of the species are gilt-head bream and red sea bream from \emph{Sparidae} family, while the other two are striped-read mullet and red mullet from \emph{Mullidae} family. Images of gilt-head bream and striped-red mullet on two types of backgrounds are used as the training and validation set. Afterward, we pick images of red sea bream and red mullet as OOD data. The Noisy FISH dataset is obtained by adding Gaussian noise to the validation set. The results for each augmentation on this dataset are listed in Table \ref{FISH_rslt}.
\begin{table}[ht]
\footnotesize
    \centering
    \begin{tabular}{|l|l|l|l|l|}
    \hline
    Training set &	$\mbox{IOU}_{val}$&$\mbox{IOU}_{noisy}$&$\mbox{IOU}_{OOD}$&SBI\\
    \hline
    No Augmentation&0.998&0.847&0.978&2.294\\
    \hline
    Color Jitter&0.996&0.884&0.990&4.014\\
    \hline
    NST&0.988&\textbf{0.932}&0.986&\textbf{4.378}\\
    \hline
    Sep. Color Jitter&0.997&0.889&\textbf{0.994}&3.882\\
    \hline
    Negative Insertion&0.865&0.644&0.895&3.924\\
    \hline
    Random Resized Crop&0.991&0.768&0.710&2.188\\
    \hline
    Random Corp Reflect&0.949&0.677&0.667&2.068\\
    \hline
    \end{tabular}
    \caption{\footnotesize Performance of models trained on datasets with different augmentations when applied to the validation set, dataset with noise injected, and dataset with naturally occurred domain shifts for FISH}.
    \vspace{-6mm}
    \label{FISH_rslt}
\end{table}

In experiments with FISH, Random Crop Reflect and Random Crop Resize result in a further decrease in the already low SBI. The rest of the augmentations unanimously caused a big increase in their respective models’ SBIs as predicted. Color Jitter, NST, and Sep. Color Jitter continues to demonstrate improved performance on Noisy and OOD data compared with the baseline. Unlike the other augmentations which increase models’ SBI, Negative Insertion shows inferior prediction accuracy even on the validation set. Though there are different augmentation methods that may theoretically encourage shape learning, not all of them will encourage the training of a segmentation network with higher performance and robustness. 

The effects of different augmentations on the training of segmentation networks on BFGT and FISH confirm the statement that when non-shape features are suppressed, shape-learning will be encouraged. It also reflects the nature of SBI as a metric measuring how much a network weighs shape in its decision process. Models with higher SBI, or tendency to learn shape, do not necessarily have higher performance on the Noisy and OOD datasets. However, models with higher performance on the Noisy and OOD datasets tend to have high SBIs.   
\begin{figure}[ht]
\centering
\begin{subfigure}[b]{0.24\textwidth}
 \centering
    \includegraphics[width=\textwidth]{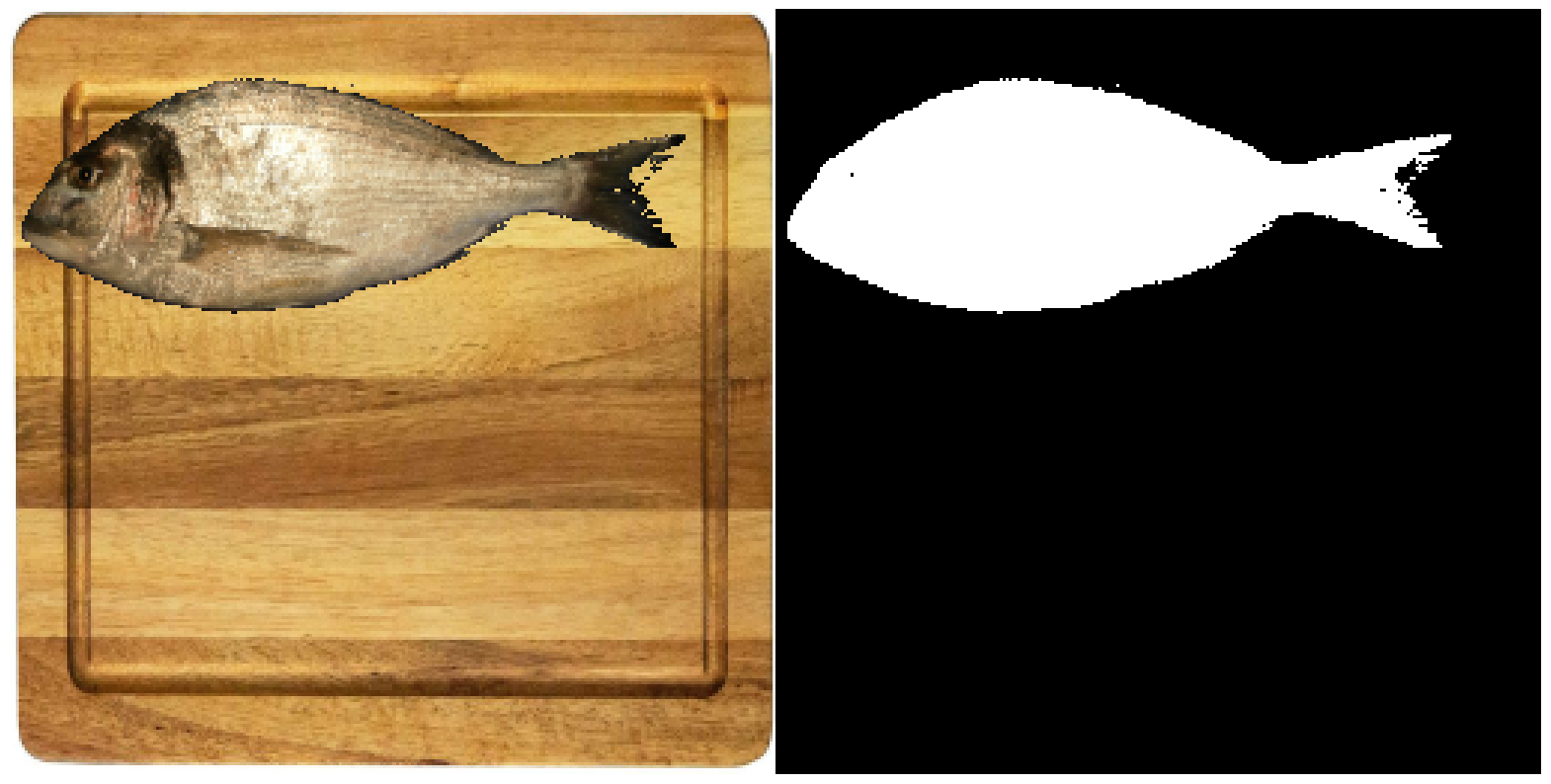}
    \caption{\scriptsize Image of gilt-head bream}
\end{subfigure}
\begin{subfigure}[b]{0.24\textwidth}
 \centering
    \includegraphics[width=\textwidth]{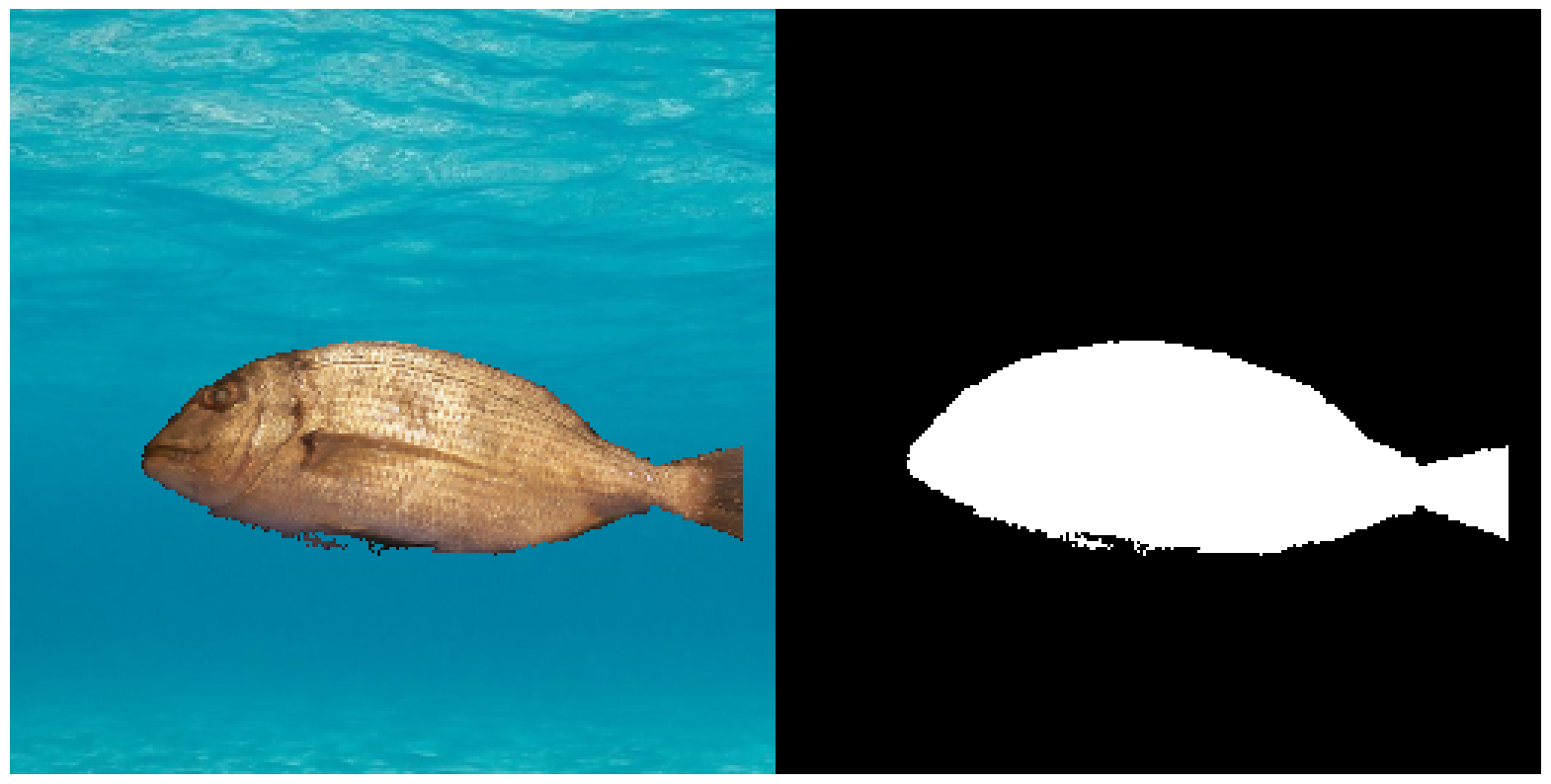}
    \caption{\scriptsize Image of red sea bream}
\end{subfigure}

\begin{subfigure}[b]{0.48\textwidth}
 \centering
    \includegraphics[width=\textwidth]{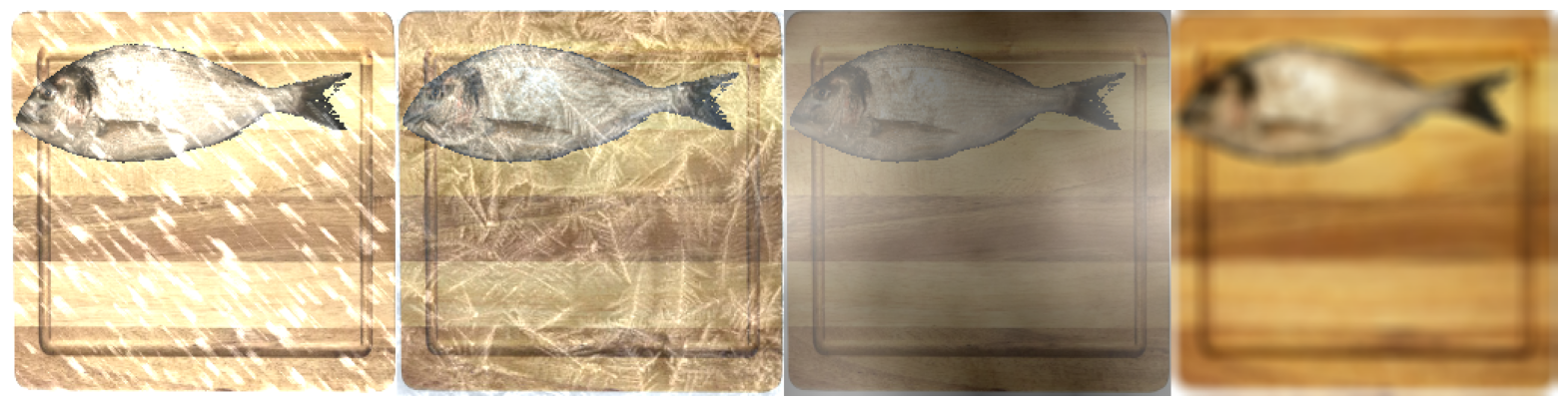}
    \caption{\scriptsize The image of gilt-head bream with (from left to right) snow filter, frost filter, fog filter, and defocus blur applied.}
\end{subfigure}
\caption{\footnotesize Samples from the three partitions of data domain in FISH}
\vspace{-1mm}
\label{FISH_samples}
\end{figure}

\subsubsection{Validation with LUNG}
The LUNG dataset is a collection of axial view lung CTs from three sources. The training and validation set consists of approximately 4000 slices extracted from 111 patient CT volumes from Pulmonary Fibrosis Progression (PFP), with their ground truth lung masks available in “CT Lung \& Heart \& Trachea segmentation” in Kaggle. We adopt another 90 axial view slices of lung CT volumes from “COVID-19 CT scans” showing lung with pneumonia as naturally occurring OOD data. (Access details in Section \ref{data}) The difference in the appearance of the COVID slices from the training set results not only from different imaging parameters but also from different infections. Each image is processed to have dimension $256\times256$. We report below the mean IOUs and SBI of models trained on different models.
\begin{table}[ht]
\footnotesize
    \centering
    \begin{tabular}{|l|l|l|l|l|}
    \hline
    Training set &	$\mbox{IOU}_{val}$&$\mbox{IOU}_{noisy}$&$\mbox{IOU}_{OOD}$&SBI\\
    \hline
    No Augmentation&0.982&0.938&\textbf{0.962}&2.650\\
    \hline
    Color Jitter&0.978&0.935&\textbf{0.962}&2.954\\
    \hline
    NST&0.978&\textbf{0.941}&\textbf{0.962}&3.014\\
    \hline
    Sep. Color Jitter&0.977&0.917&0.956&2.650\\
    \hline
    Negative Insertion&0.981&0.931&0.953&\textbf{3.022}\\
    \hline
    Random Resized Crop&0.942&0.793&0.853&2.534\\
    \hline
    Random Crop Reflect&0.979&0.672&0.943&2.128\\
    \hline
    \end{tabular}
    \caption{\footnotesize Performance of models trained on datasets with different augmentations when applied to the validation set, dataset with noise injected, and dataset with naturally occurred domain shifts for LUNG}.
    \vspace{-2mm}
    \label{LUNG_rslt}
\end{table}
Unlike in BFGT and FISH, as shown in Table \ref{LUNG_rslt}, augmentations have much smaller effects on the performance of LUNG models. All augmentations that preserve targets’ shape yield models with very close performance on Validation, Noisy, and OOD data. Despite this, the effects of augmentations on SBI still align with our assumptions. Again, Random Resized Crop and Random Crop Reflect yield models with smaller SBI than the baseline, while the remaining augmentations increase each model’s SBI. 
\begin{figure}[ht]
\centering
\begin{subfigure}[b]{0.24\textwidth}
 \centering
    \includegraphics[width=\textwidth]{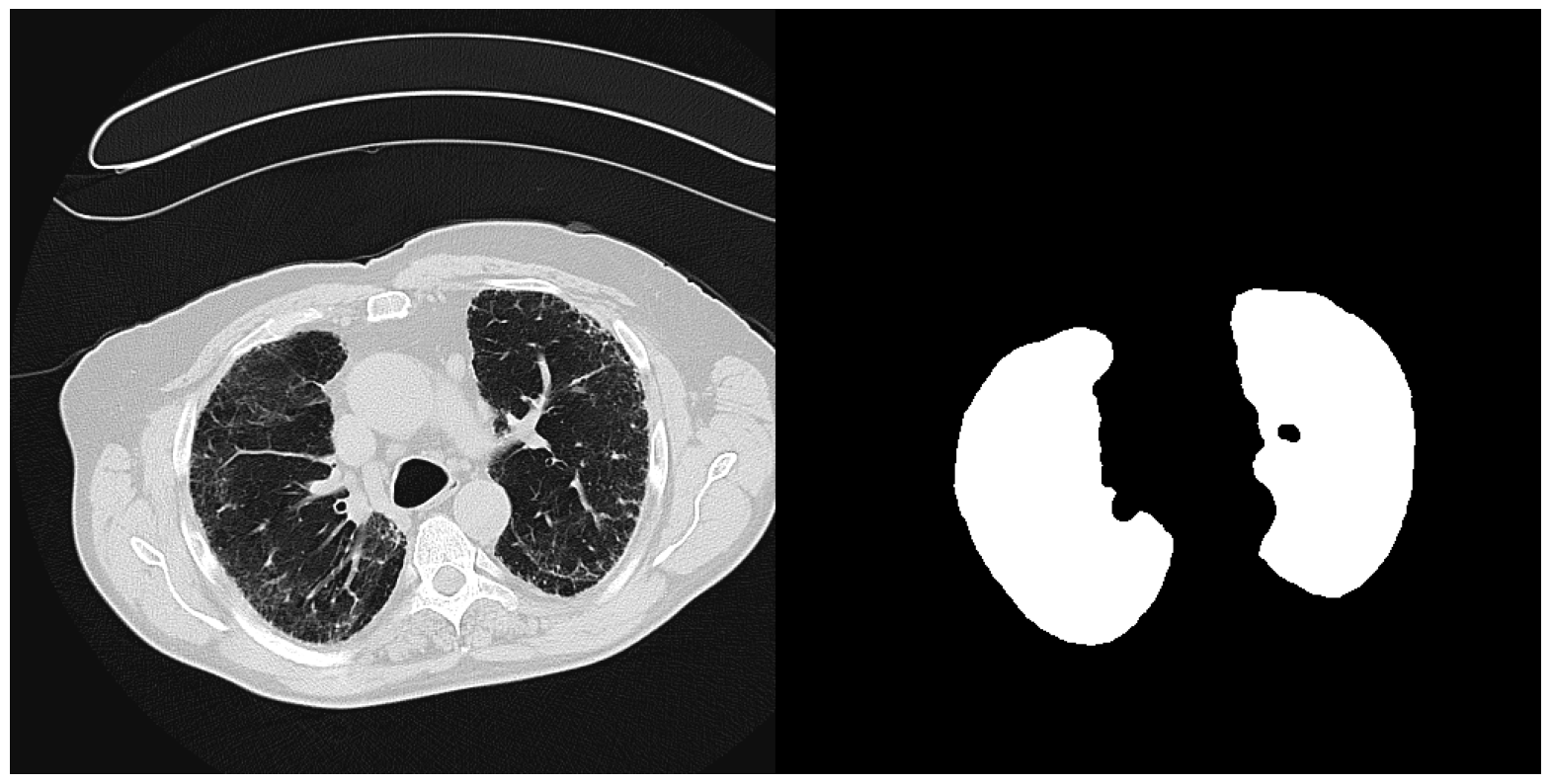}
    \caption{\scriptsize Image from "OSIC Pulmonary Fibrosis Progression"~\cite{OSIC_PFP,OSIC_PFP_msk}}
\end{subfigure}
\begin{subfigure}[b]{0.24\textwidth}
 \centering
    \includegraphics[width=\textwidth]{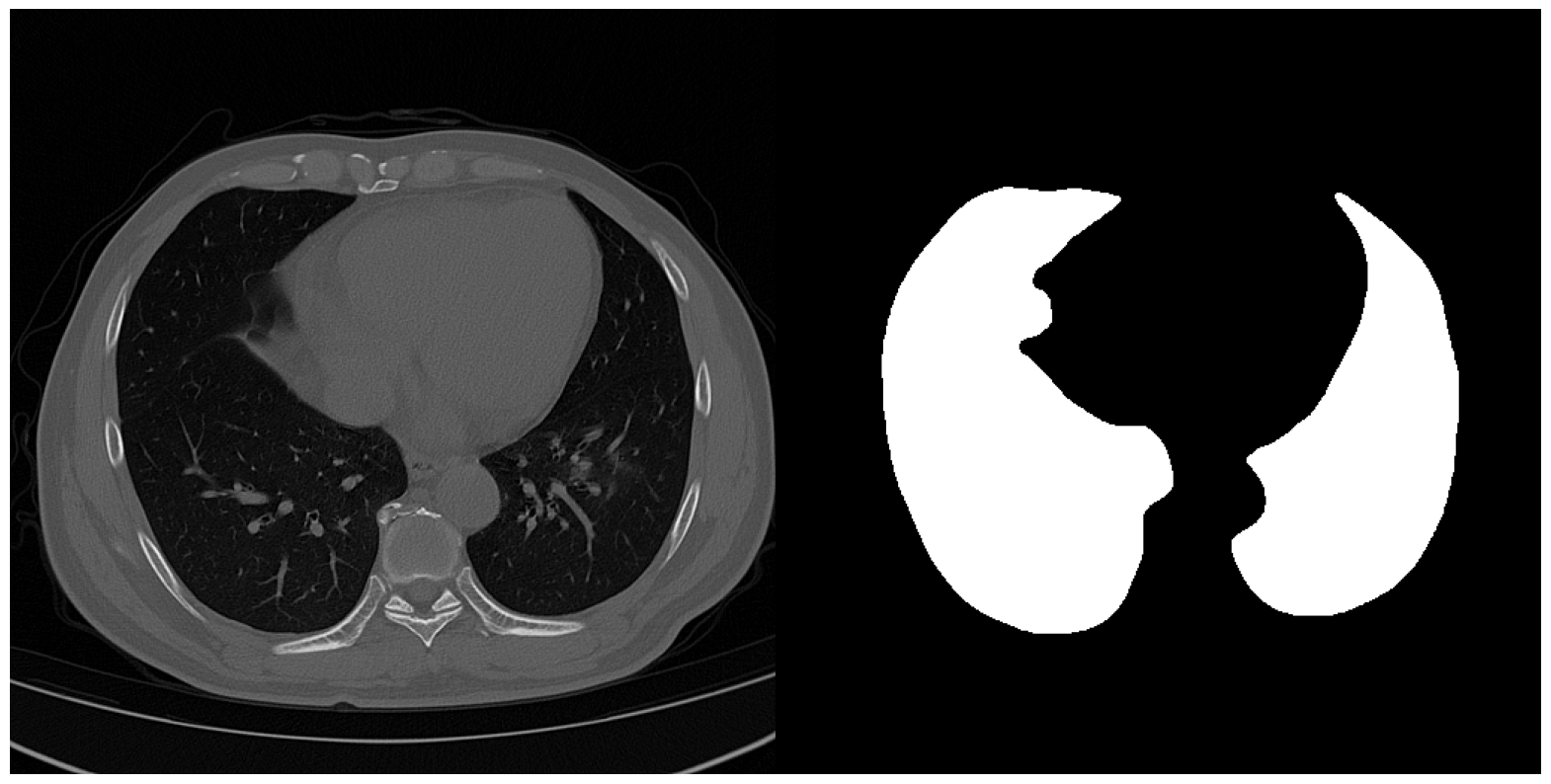}
    \caption{\scriptsize Image from "COVID-19 CT scans"~\cite{glick,pavia,ma_jun_2020_3757476}}
\end{subfigure}

\begin{subfigure}[b]{0.48\textwidth}
 \centering
    \includegraphics[width=\textwidth]{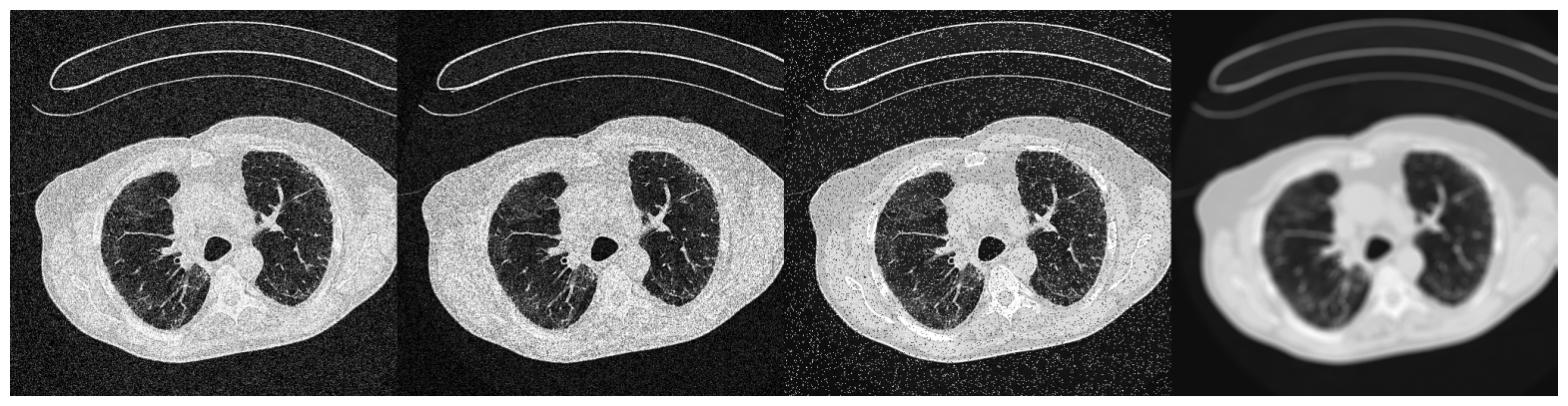}
    \caption{\scriptsize The image from "OSIC Pulmonary Fibrosis Progression" with (from left to right) Gaussian noise, shot noise, impulse noise, and defocus blur applied.}
\end{subfigure}
\caption{\footnotesize Samples from the three partitions of data domain in FISH}
\vspace{-1mm}
\label{LUNG_samples}
\end{figure}

Augmentations, however, have negligible effects on this dataset. We explain this phenomenon by arguing that non-shape features may persist across multiple domains for LUNG. In LUNG, pixels associated with the lung have a texture style easily distinguishable from those associated with the rest of the image. The cooccurrence of high-intensity contours formed by cartilages also helps locate the lung. These features persist even across different image domains. In this case, the shape may not be the best discriminative feature in prediction capability. Non-shape features may be the more desirable feature sets to learn when optimizing the Cross-Entropy objective function. To avoid overstretching on interpreting the results, we will stop here with the conclusion that though Color Jitter and NST may suppress non-shape features, they barely hurt a model’s generalizability on Noisy and OOD data. 

On all three datasets, we show that suppressing non-shape features encourages shape learning. Our explorations made on synthetic data are hence generalizable to many real datasets. Color Jitter and NST provide superior augmentation performance on both noisy and OOD data but are computationally intensive. Compared to NST, Color Jitter is a faster option but yields a smaller performance boost. 

\section{Conclusion\&Discussion}
\label{con}
In this paper, we defined shape and designed a behavioral metric that measures a segmentation network’s tendency to consider shape by measuring its response to variations in input features. Given the definition of shape and our model-independent metric, we designed an intuitive yet comprehensive experiment to explore whether, and under what circumstances, a network can utilize shape as a discriminative feature. 

To the best of our knowledge, our experiment is the first of its type that revealed shape to be the least prioritized feature learned by CNNs. Learning shape information requires the absence of alternative correlated features. To obtain a shape-learning CNN, strong regularization or augmentation would be needed in practice. Furthermore, a sufficiently large receptive field size relative to object size is vital for a segmentation network to learn shape. We also concluded that shape learning is indeed a useful property of a network as it helps with out-of-domain generalization. We confirm our results from synthetic data on reproducible real-life scenarios and showed that Color Jitter and NST can be used as augmentations for encouraging shape learning. 

Our study had some limitations. First, our study assumed the objects of interest to have identical, or at least similar shapes which may not hold in some visual tasks in natural images.  While this is a limitation, this assumption allowed for a systematic evaluation of the issue at hand. Furthermore, as stated in Section \ref{SBI}, SBI is an unnormalized metric. It does not allow for comparisons across models trained for different tasks. One also needs to apply care in properly selecting 
$D_{rm}$(e.g., non-shape information is sufficiently removed) since otherwise it may lead to a misleading SBI that doesn’t reflect the true tendency of a model to consider shape in the inference process.

Some future studies could be conducted on models beyond those consisting only of segmentation networks. Some examples include those that embed convolutional layers with graphical neural networks and attention mechanisms, or even those that use only non-convolutional approaches such as models derived from visual transformers. These models intuitively could have a better tendency to extract high-level geometric features than traditional CNNs given their lack of global translational invariance. It may also be beneficial to analyze the effects of different loss functions and training techniques \cite{NEURIPS2022_fb64a552}, especially those loss functions with assumptions on shape priors. These studies have the potential to bring together a more complete understanding of the feature learning tendency of deep neural networks for object recognition tasks.

Still, one should remember that encouraging shape learning is not always beneficial. In some settings, it is the color or the texture the most reliable feature. An example of such a scenario is the segmentation of brain tumors, where tumors can be easily identified by applying intensity thresholds to the different MRI sequences. In such situations, encouraging shape learning through data augmentation may lower a model’s performance even in the presence of potential OOD data distribution. 

\section*{Acknowledgments}
We would like to express our gratitude to Gregory Szumel, Nicholas Konz, and Hanxue Gu from Mazurowski Lab for their suggestions on the readability of this paper. 
\bibliographystyle{elsarticle-num-names} 


\bibliography{bib.bib}

\begin{thebibliography}{47}
\expandafter\ifx\csname natexlab\endcsname\relax\def\natexlab#1{#1}\fi
\providecommand{\url}[1]{\texttt{#1}}
\providecommand{\href}[2]{#2}
\providecommand{\path}[1]{#1}
\providecommand{\DOIprefix}{doi:}
\providecommand{\ArXivprefix}{arXiv:}
\providecommand{\URLprefix}{URL: }
\providecommand{\Pubmedprefix}{pmid:}
\providecommand{\doi}[1]{\href{http://dx.doi.org/#1}{\path{#1}}}
\providecommand{\Pubmed}[1]{\href{pmid:#1}{\path{#1}}}
\providecommand{\bibinfo}[2]{#2}
\ifx\xfnm\relax \def\xfnm[#1]{\unskip,\space#1}\fi
%Type = Article
\bibitem[{Shen et~al.(2015)Shen, Deng, Cheng, Nie, and Yan}]{shen2015robust}
\bibinfo{author}{J.~Shen}, \bibinfo{author}{R.~H. Deng}, \bibinfo{author}{Z.~Cheng}, \bibinfo{author}{L.~Nie}, \bibinfo{author}{S.~Yan},
\newblock \bibinfo{title}{On robust image spam filtering via comprehensive visual modeling},
\newblock \bibinfo{journal}{Pattern Recognition} \bibinfo{volume}{48} (\bibinfo{year}{2015}) \bibinfo{pages}{3227--3238}.
%Type = Article
\bibitem[{Hong et~al.(2021)Hong, Gao, Yokoya, Yao, Chanussot, Du, and Zhang}]{9174822}
\bibinfo{author}{D.~Hong}, \bibinfo{author}{L.~Gao}, \bibinfo{author}{N.~Yokoya}, \bibinfo{author}{J.~Yao}, \bibinfo{author}{J.~Chanussot}, \bibinfo{author}{Q.~Du}, \bibinfo{author}{B.~Zhang},
\newblock \bibinfo{title}{More diverse means better: Multimodal deep learning meets remote-sensing imagery classification},
\newblock \bibinfo{journal}{IEEE Transactions on Geoscience and Remote Sensing} \bibinfo{volume}{59} (\bibinfo{year}{2021}) \bibinfo{pages}{4340--4354}. \DOIprefix\doi{10.1109/TGRS.2020.3016820}.
%Type = Article
\bibitem[{Wu et~al.(2022)Wu, Hong, and Chanussot}]{9598903}
\bibinfo{author}{X.~Wu}, \bibinfo{author}{D.~Hong}, \bibinfo{author}{J.~Chanussot},
\newblock \bibinfo{title}{Convolutional neural networks for multimodal remote sensing data classification},
\newblock \bibinfo{journal}{IEEE Transactions on Geoscience and Remote Sensing} \bibinfo{volume}{60} (\bibinfo{year}{2022}) \bibinfo{pages}{1--10}. \DOIprefix\doi{10.1109/TGRS.2021.3124913}.
%Type = Article
\bibitem[{Anand et~al.(2023)Anand, Gupta, Koundal, and Singh}]{anand2023fusion}
\bibinfo{author}{V.~Anand}, \bibinfo{author}{S.~Gupta}, \bibinfo{author}{D.~Koundal}, \bibinfo{author}{K.~Singh},
\newblock \bibinfo{title}{Fusion of u-net and cnn model for segmentation and classification of skin lesion from dermoscopy images},
\newblock \bibinfo{journal}{Expert Systems with Applications} \bibinfo{volume}{213} (\bibinfo{year}{2023}) \bibinfo{pages}{119230}.
%Type = Inproceedings
\bibitem[{Zhao et~al.(2017)Zhao, Shi, Qi, Wang, and Jia}]{zhao2017pyramid}
\bibinfo{author}{H.~Zhao}, \bibinfo{author}{J.~Shi}, \bibinfo{author}{X.~Qi}, \bibinfo{author}{X.~Wang}, \bibinfo{author}{J.~Jia},
\newblock \bibinfo{title}{Pyramid scene parsing network},
\newblock in: \bibinfo{booktitle}{Proceedings of the IEEE conference on computer vision and pattern recognition}, \bibinfo{year}{2017}, pp. \bibinfo{pages}{2881--2890}.
%Type = Inproceedings
\bibitem[{Chen et~al.(2018)Chen, Zhu, Papandreou, Schroff, and Adam}]{Chen_2018_ECCV}
\bibinfo{author}{L.-C. Chen}, \bibinfo{author}{Y.~Zhu}, \bibinfo{author}{G.~Papandreou}, \bibinfo{author}{F.~Schroff}, \bibinfo{author}{H.~Adam},
\newblock \bibinfo{title}{Encoder-decoder with atrous separable convolution for semantic image segmentation},
\newblock in: \bibinfo{booktitle}{Proceedings of the European Conference on Computer Vision (ECCV)}, \bibinfo{year}{2018}.
%Type = Inproceedings
\bibitem[{Liu et~al.(2022)Liu, Mao, Wu, Feichtenhofer, Darrell, and Xie}]{liu2022convnet}
\bibinfo{author}{Z.~Liu}, \bibinfo{author}{H.~Mao}, \bibinfo{author}{C.-Y. Wu}, \bibinfo{author}{C.~Feichtenhofer}, \bibinfo{author}{T.~Darrell}, \bibinfo{author}{S.~Xie},
\newblock \bibinfo{title}{A convnet for the 2020s},
\newblock in: \bibinfo{booktitle}{Proceedings of the IEEE/CVF conference on computer vision and pattern recognition}, \bibinfo{year}{2022}, pp. \bibinfo{pages}{11976--11986}.
%Type = Article
\bibitem[{Kriegeskorte(2015)}]{kriegeskorte2015deep}
\bibinfo{author}{N.~Kriegeskorte},
\newblock \bibinfo{title}{Deep neural networks: a new framework for modeling biological vision and brain information processing},
\newblock \bibinfo{journal}{Annual review of vision science} \bibinfo{volume}{1} (\bibinfo{year}{2015}) \bibinfo{pages}{417--446}.
%Type = Article
\bibitem[{LeCun et~al.(2015)LeCun, Bengio, and Hinton}]{lecun2015deep}
\bibinfo{author}{Y.~LeCun}, \bibinfo{author}{Y.~Bengio}, \bibinfo{author}{G.~Hinton},
\newblock \bibinfo{title}{Deep learning},
\newblock \bibinfo{journal}{nature} \bibinfo{volume}{521} (\bibinfo{year}{2015}) \bibinfo{pages}{436--444}.
%Type = Inproceedings
\bibitem[{Geirhos et~al.(2019)Geirhos, Rubisch, Michaelis, Bethge, Wichmann, and Brendel}]{geirhos2018imagenet}
\bibinfo{author}{R.~Geirhos}, \bibinfo{author}{P.~Rubisch}, \bibinfo{author}{C.~Michaelis}, \bibinfo{author}{M.~Bethge}, \bibinfo{author}{F.~A. Wichmann}, \bibinfo{author}{W.~Brendel},
\newblock \bibinfo{title}{Imagenet-trained cnns are biased towards texture; increasing shape bias improves accuracy and robustness},
\newblock in: \bibinfo{booktitle}{7th International Conference on Learning Representations, {ICLR} 2019, New Orleans, LA, USA, May 6-9, 2019}, \bibinfo{year}{2019}.
%Type = Article
\bibitem[{Wichmann and Geirhos(2023)}]{doi:10.1146/annurev-vision-120522-031739}
\bibinfo{author}{F.~A. Wichmann}, \bibinfo{author}{R.~Geirhos},
\newblock \bibinfo{title}{Are deep neural networks adequate behavioral models of human visual perception?},
\newblock \bibinfo{journal}{Annual Review of Vision Science} \bibinfo{volume}{9} (\bibinfo{year}{2023}) \bibinfo{pages}{null}. \DOIprefix\doi{10.1146/annurev-vision-120522-031739}, \bibinfo{note}{pMID: 37001509}.
%Type = Inproceedings
\bibitem[{Mummadi et~al.(2021)Mummadi, Subramaniam, Hutmacher, Vitay, Fischer, and Metzen}]{mummadi2021does}
\bibinfo{author}{C.~K. Mummadi}, \bibinfo{author}{R.~Subramaniam}, \bibinfo{author}{R.~Hutmacher}, \bibinfo{author}{J.~Vitay}, \bibinfo{author}{V.~Fischer}, \bibinfo{author}{J.~H. Metzen},
\newblock \bibinfo{title}{Does enhanced shape bias improve neural network robustness to common corruptions?},
\newblock in: \bibinfo{booktitle}{9th International Conference on Learning Representations, {ICLR} 2021, Virtual Event, Austria, May 3-7, 2021}, \bibinfo{year}{2021}.
%Type = Inproceedings
\bibitem[{Tram{\`{e}}r et~al.(2018)Tram{\`{e}}r, Kurakin, Papernot, Goodfellow, Boneh, and McDaniel}]{DBLP:conf/iclr/TramerKPGBM18}
\bibinfo{author}{F.~Tram{\`{e}}r}, \bibinfo{author}{A.~Kurakin}, \bibinfo{author}{N.~Papernot}, \bibinfo{author}{I.~J. Goodfellow}, \bibinfo{author}{D.~Boneh}, \bibinfo{author}{P.~D. McDaniel},
\newblock \bibinfo{title}{Ensemble adversarial training: Attacks and defenses},
\newblock in: \bibinfo{booktitle}{6th International Conference on Learning Representations, {ICLR} 2018, Vancouver, BC, Canada, April 30 - May 3, 2018, Conference Track Proceedings}, \bibinfo{publisher}{OpenReview.net}, \bibinfo{year}{2018}. \URLprefix \url{https://openreview.net/forum?id=rkZvSe-RZ}.
%Type = Article
\bibitem[{Shen and Robertson(2021)}]{shen2021bbas}
\bibinfo{author}{J.~Shen}, \bibinfo{author}{N.~Robertson},
\newblock \bibinfo{title}{Bbas: Towards large scale effective ensemble adversarial attacks against deep neural network learning},
\newblock \bibinfo{journal}{Information Sciences} \bibinfo{volume}{569} (\bibinfo{year}{2021}) \bibinfo{pages}{469--478}.
%Type = Inproceedings
\bibitem[{Cohen et~al.(2019)Cohen, Rosenfeld, and Kolter}]{pmlr-v97-cohen19c}
\bibinfo{author}{J.~Cohen}, \bibinfo{author}{E.~Rosenfeld}, \bibinfo{author}{Z.~Kolter},
\newblock \bibinfo{title}{Certified adversarial robustness via randomized smoothing},
\newblock in: \bibinfo{editor}{K.~Chaudhuri}, \bibinfo{editor}{R.~Salakhutdinov} (Eds.), \bibinfo{booktitle}{Proceedings of the 36th International Conference on Machine Learning}, volume~\bibinfo{volume}{97} of \textit{\bibinfo{series}{Proceedings of Machine Learning Research}}, \bibinfo{publisher}{PMLR}, \bibinfo{year}{2019}, pp. \bibinfo{pages}{1310--1320}. \URLprefix \url{https://proceedings.mlr.press/v97/cohen19c.html}.
%Type = Article
\bibitem[{Shafahi et~al.(2019)Shafahi, Najibi, Ghiasi, Xu, Dickerson, Studer, Davis, Taylor, and Goldstein}]{shafahi2019adversarial}
\bibinfo{author}{A.~Shafahi}, \bibinfo{author}{M.~Najibi}, \bibinfo{author}{M.~A. Ghiasi}, \bibinfo{author}{Z.~Xu}, \bibinfo{author}{J.~Dickerson}, \bibinfo{author}{C.~Studer}, \bibinfo{author}{L.~S. Davis}, \bibinfo{author}{G.~Taylor}, \bibinfo{author}{T.~Goldstein},
\newblock \bibinfo{title}{Adversarial training for free!},
\newblock \bibinfo{journal}{Advances in Neural Information Processing Systems} \bibinfo{volume}{32} (\bibinfo{year}{2019}).
%Type = Article
\bibitem[{Ilyas et~al.(2019)Ilyas, Santurkar, Tsipras, Engstrom, Tran, and Madry}]{ilyas2019adversarial}
\bibinfo{author}{A.~Ilyas}, \bibinfo{author}{S.~Santurkar}, \bibinfo{author}{D.~Tsipras}, \bibinfo{author}{L.~Engstrom}, \bibinfo{author}{B.~Tran}, \bibinfo{author}{A.~Madry},
\newblock \bibinfo{title}{Adversarial examples are not bugs, they are features},
\newblock \bibinfo{journal}{Advances in neural information processing systems} \bibinfo{volume}{32} (\bibinfo{year}{2019}).
%Type = Article
\bibitem[{Kubilius et~al.(2016)Kubilius, Bracci, and Op~de Beeck}]{kubilius2016deep}
\bibinfo{author}{J.~Kubilius}, \bibinfo{author}{S.~Bracci}, \bibinfo{author}{H.~P. Op~de Beeck},
\newblock \bibinfo{title}{Deep neural networks as a computational model for human shape sensitivity},
\newblock \bibinfo{journal}{PLoS computational biology} \bibinfo{volume}{12} (\bibinfo{year}{2016}) \bibinfo{pages}{e1004896}.
%Type = Article
\bibitem[{Baker et~al.(2018)Baker, Lu, Erlikhman, and Kellman}]{baker2018deep}
\bibinfo{author}{N.~Baker}, \bibinfo{author}{H.~Lu}, \bibinfo{author}{G.~Erlikhman}, \bibinfo{author}{P.~J. Kellman},
\newblock \bibinfo{title}{Deep convolutional networks do not classify based on global object shape},
\newblock \bibinfo{journal}{PLoS computational biology} \bibinfo{volume}{14} (\bibinfo{year}{2018}) \bibinfo{pages}{e1006613}.
%Type = Article
\bibitem[{Baker et~al.(2020)Baker, Lu, Erlikhman, and Kellman}]{baker2020local}
\bibinfo{author}{N.~Baker}, \bibinfo{author}{H.~Lu}, \bibinfo{author}{G.~Erlikhman}, \bibinfo{author}{P.~J. Kellman},
\newblock \bibinfo{title}{Local features and global shape information in object classification by deep convolutional neural networks},
\newblock \bibinfo{journal}{Vision research} \bibinfo{volume}{172} (\bibinfo{year}{2020}) \bibinfo{pages}{46--61}.
%Type = Inproceedings
\bibitem[{Ritter et~al.(2017)Ritter, Barrett, Santoro, and Botvinick}]{ritter2017cognitive}
\bibinfo{author}{S.~Ritter}, \bibinfo{author}{D.~G. Barrett}, \bibinfo{author}{A.~Santoro}, \bibinfo{author}{M.~M. Botvinick},
\newblock \bibinfo{title}{Cognitive psychology for deep neural networks: A shape bias case study},
\newblock in: \bibinfo{booktitle}{International conference on machine learning}, \bibinfo{organization}{PMLR}, \bibinfo{year}{2017}, pp. \bibinfo{pages}{2940--2949}.
%Type = Article
\bibitem[{Vinyals et~al.(2016)Vinyals, Blundell, Lillicrap, Wierstra et~al.}]{vinyals2016matching}
\bibinfo{author}{O.~Vinyals}, \bibinfo{author}{C.~Blundell}, \bibinfo{author}{T.~Lillicrap}, \bibinfo{author}{D.~Wierstra}, et~al.,
\newblock \bibinfo{title}{Matching networks for one shot learning},
\newblock \bibinfo{journal}{Advances in neural information processing systems} \bibinfo{volume}{29} (\bibinfo{year}{2016}).
%Type = Article
\bibitem[{Hermann et~al.(2020)Hermann, Chen, and Kornblith}]{hermann2020origins}
\bibinfo{author}{K.~Hermann}, \bibinfo{author}{T.~Chen}, \bibinfo{author}{S.~Kornblith},
\newblock \bibinfo{title}{The origins and prevalence of texture bias in convolutional neural networks},
\newblock \bibinfo{journal}{Advances in Neural Information Processing Systems} \bibinfo{volume}{33} (\bibinfo{year}{2020}) \bibinfo{pages}{19000--19015}.
%Type = Article
\bibitem[{Hermann and Lampinen(2020)}]{hermann2020shapes}
\bibinfo{author}{K.~Hermann}, \bibinfo{author}{A.~Lampinen},
\newblock \bibinfo{title}{What shapes feature representations? exploring datasets, architectures, and training},
\newblock \bibinfo{journal}{Advances in Neural Information Processing Systems} \bibinfo{volume}{33} (\bibinfo{year}{2020}) \bibinfo{pages}{9995--10006}.
%Type = Inproceedings
\bibitem[{Hosseini et~al.(2018)Hosseini, Xiao, Jaiswal, and Poovendran}]{hosseini2018assessing}
\bibinfo{author}{H.~Hosseini}, \bibinfo{author}{B.~Xiao}, \bibinfo{author}{M.~Jaiswal}, \bibinfo{author}{R.~Poovendran},
\newblock \bibinfo{title}{Assessing shape bias property of convolutional neural networks},
\newblock in: \bibinfo{booktitle}{Proceedings of the IEEE Conference on Computer Vision and Pattern Recognition Workshops}, \bibinfo{year}{2018}, pp. \bibinfo{pages}{1923--1931}.
%Type = Inproceedings
\bibitem[{Li et~al.(2021)Li, Yu, Tan, Mei, Tang, Shen, Yuille, and Xie}]{li2020shape}
\bibinfo{author}{Y.~Li}, \bibinfo{author}{Q.~Yu}, \bibinfo{author}{M.~Tan}, \bibinfo{author}{J.~Mei}, \bibinfo{author}{P.~Tang}, \bibinfo{author}{W.~Shen}, \bibinfo{author}{A.~L. Yuille}, \bibinfo{author}{C.~Xie},
\newblock \bibinfo{title}{Shape-texture debiased neural network training},
\newblock in: \bibinfo{booktitle}{9th International Conference on Learning Representations, {ICLR} 2021, Virtual Event, Austria, May 3-7, 2021}, \bibinfo{year}{2021}.
%Type = Inproceedings
\bibitem[{Hendrycks and Dietterich(2019)}]{hendrycks2019benchmarking}
\bibinfo{author}{D.~Hendrycks}, \bibinfo{author}{T.~G. Dietterich},
\newblock \bibinfo{title}{Benchmarking neural network robustness to common corruptions and perturbations},
\newblock in: \bibinfo{booktitle}{7th International Conference on Learning Representations, {ICLR} 2019, New Orleans, LA, USA, May 6-9, 2019}, \bibinfo{year}{2019}.
%Type = Inproceedings
\bibitem[{Islam et~al.(2021)Islam, Kowal, Esser, Jia, Ommer, Derpanis, and Bruce}]{Islam2021ShapeOT}
\bibinfo{author}{M.~A. Islam}, \bibinfo{author}{M.~Kowal}, \bibinfo{author}{P.~Esser}, \bibinfo{author}{S.~Jia}, \bibinfo{author}{B.~Ommer}, \bibinfo{author}{K.~G. Derpanis}, \bibinfo{author}{N.~D.~B. Bruce},
\newblock \bibinfo{title}{Shape or texture: Understanding discriminative features in cnns},
\newblock in: \bibinfo{booktitle}{9th International Conference on Learning Representations, {ICLR} 2021, Virtual Event, Austria, May 3-7, 2021}, \bibinfo{year}{2021}.
%Type = Inproceedings
\bibitem[{Tripathi et~al.(2023)Tripathi, Singh, Chakraborty, and Shenoy}]{Tripathi_2023_CVPR}
\bibinfo{author}{A.~Tripathi}, \bibinfo{author}{R.~Singh}, \bibinfo{author}{A.~Chakraborty}, \bibinfo{author}{P.~Shenoy},
\newblock \bibinfo{title}{Edges to shapes to concepts: Adversarial augmentation for robust vision},
\newblock in: \bibinfo{booktitle}{Proceedings of the IEEE/CVF Conference on Computer Vision and Pattern Recognition (CVPR)}, \bibinfo{year}{2023}, pp. \bibinfo{pages}{24470--24479}.
%Type = Inproceedings
\bibitem[{Sinha and Kovashka(2023)}]{sinha2023towards}
\bibinfo{author}{H.~Sinha}, \bibinfo{author}{A.~Kovashka},
\newblock \bibinfo{title}{Towards shape-regularized learning for mitigating texture bias in cnns},
\newblock in: \bibinfo{booktitle}{Proceedings of the 2023 ACM International Conference on Multimedia Retrieval}, \bibinfo{year}{2023}, pp. \bibinfo{pages}{325--334}.
%Type = Inproceedings
\bibitem[{Ding et~al.(2022)Ding, Zhang, Han, and Ding}]{ding2022scaling}
\bibinfo{author}{X.~Ding}, \bibinfo{author}{X.~Zhang}, \bibinfo{author}{J.~Han}, \bibinfo{author}{G.~Ding},
\newblock \bibinfo{title}{Scaling up your kernels to 31x31: Revisiting large kernel design in cnns},
\newblock in: \bibinfo{booktitle}{Proceedings of the IEEE/CVF conference on computer vision and pattern recognition}, \bibinfo{year}{2022}, pp. \bibinfo{pages}{11963--11975}.
%Type = Article
\bibitem[{Haralick et~al.(1973)Haralick, Shanmugam, and Dinstein}]{haralick1973textural}
\bibinfo{author}{R.~M. Haralick}, \bibinfo{author}{K.~Shanmugam}, \bibinfo{author}{I.~H. Dinstein},
\newblock \bibinfo{title}{Textural features for image classification},
\newblock \bibinfo{journal}{IEEE Transactions on systems, man, and cybernetics}  (\bibinfo{year}{1973}) \bibinfo{pages}{610--621}.
%Type = Article
\bibitem[{Van~Gool et~al.(1985)Van~Gool, Dewaele, and Oosterlinck}]{van1985texture}
\bibinfo{author}{L.~Van~Gool}, \bibinfo{author}{P.~Dewaele}, \bibinfo{author}{A.~Oosterlinck},
\newblock \bibinfo{title}{Texture analysis anno 1983},
\newblock \bibinfo{journal}{Computer vision, graphics, and image processing} \bibinfo{volume}{29} (\bibinfo{year}{1985}) \bibinfo{pages}{336--357}.
%Type = Article
\bibitem[{Diesendruck and Bloom(2003)}]{diesendruck2003specific}
\bibinfo{author}{G.~Diesendruck}, \bibinfo{author}{P.~Bloom},
\newblock \bibinfo{title}{How specific is the shape bias?},
\newblock \bibinfo{journal}{Child development} \bibinfo{volume}{74} (\bibinfo{year}{2003}) \bibinfo{pages}{168--178}.
%Type = Inproceedings
\bibitem[{Gatys et~al.(2017)Gatys, Ecker, Bethge, Hertzmann, and Shechtman}]{gatys2017controlling}
\bibinfo{author}{L.~A. Gatys}, \bibinfo{author}{A.~S. Ecker}, \bibinfo{author}{M.~Bethge}, \bibinfo{author}{A.~Hertzmann}, \bibinfo{author}{E.~Shechtman},
\newblock \bibinfo{title}{Controlling perceptual factors in neural style transfer},
\newblock in: \bibinfo{booktitle}{Proceedings of the IEEE conference on computer vision and pattern recognition}, \bibinfo{year}{2017}, pp. \bibinfo{pages}{3985--3993}.
%Type = Article
\bibitem[{Buda et~al.(2021)Buda, Saha, Walsh, Ghate, Li, Święcicki, Lo, and Mazurowski}]{buda2020detection}
\bibinfo{author}{M.~Buda}, \bibinfo{author}{A.~Saha}, \bibinfo{author}{R.~Walsh}, \bibinfo{author}{S.~Ghate}, \bibinfo{author}{N.~Li}, \bibinfo{author}{A.~Święcicki}, \bibinfo{author}{J.~Y. Lo}, \bibinfo{author}{M.~A. Mazurowski},
\newblock \bibinfo{title}{{A Data Set and Deep Learning Algorithm for the Detection of Masses and Architectural Distortions in Digital Breast Tomosynthesis Images}},
\newblock \bibinfo{journal}{JAMA Network Open} \bibinfo{volume}{4} (\bibinfo{year}{2021}) \bibinfo{pages}{e2119100--e2119100}. \DOIprefix\doi{10.1001/jamanetworkopen.2021.19100}.
%Type = Inproceedings
\bibitem[{Ulucan et~al.(2020)Ulucan, Karakaya, and Turkan}]{ulucan2020large}
\bibinfo{author}{O.~Ulucan}, \bibinfo{author}{D.~Karakaya}, \bibinfo{author}{M.~Turkan},
\newblock \bibinfo{title}{A large-scale dataset for fish segmentation and classification},
\newblock in: \bibinfo{booktitle}{2020 Innovations in Intelligent Systems and Applications Conference (ASYU)}, \bibinfo{organization}{IEEE}, \bibinfo{year}{2020}, pp. \bibinfo{pages}{1--5}.
%Type = Techreport
\bibitem[{{Open Source Imaging Consortium (OSIC)}(2020)}]{OSIC_PFP}
\bibinfo{author}{{Open Source Imaging Consortium (OSIC)}}, \bibinfo{title}{{OSIC Pulmonary Fibrosis Progression}}, \bibinfo{type}{Technical Report}, \bibinfo{year}{2020}. \URLprefix \url{https://www.kaggle.com/competitions/osic-pulmonary-fibrosis-progression/}.
%Type = Techreport
\bibitem[{{Kónya} et~al.(2020)}]{OSIC_PFP_msk}
\bibinfo{author}{{Kónya}}, et~al., \bibinfo{title}{{CT Lung \& Heart \& Trachea segmentation}}, \bibinfo{type}{Technical Report}, \bibinfo{year}{2020}. \URLprefix \url{https://www.kaggle.com/sandorkonya/ct-lung-heart-trachea-segmentation}.
%Type = Techreport
\bibitem[{Glick(2020)}]{glick}
\bibinfo{author}{Y.~Glick}, \bibinfo{title}{{Viewing Playlist: COVID-19 Pneumonia | Radiopaedia.Org}}, \bibinfo{type}{Technical Report}, \bibinfo{year}{2020}. \URLprefix \url{https://radiopaedia.org/playlists/25887}.
%Type = Techreport
\bibitem[{Paiva(2020)}]{pavia}
\bibinfo{author}{O.~Paiva}, \bibinfo{title}{{Helping Radiologists To Help People In More Than 100 Countries!| Coronavirus Cases}}, \bibinfo{type}{Technical Report}, \bibinfo{year}{2020}. \URLprefix \url{coronacases.org}.
%Type = Techreport
\bibitem[{Ma et~al.(2020)Ma, Ge, Wang, An, Gao, Yu, Zhang, Liu, Deng, Cao, Wei, Mei, Yang, Nie, Li, Tian, Zhu, Zhu, Dong, and He}]{ma_jun_2020_3757476}
\bibinfo{author}{J.~Ma}, \bibinfo{author}{C.~Ge}, \bibinfo{author}{Y.~Wang}, \bibinfo{author}{X.~An}, \bibinfo{author}{J.~Gao}, \bibinfo{author}{Z.~Yu}, \bibinfo{author}{M.~Zhang}, \bibinfo{author}{X.~Liu}, \bibinfo{author}{X.~Deng}, \bibinfo{author}{S.~Cao}, \bibinfo{author}{H.~Wei}, \bibinfo{author}{S.~Mei}, \bibinfo{author}{X.~Yang}, \bibinfo{author}{Z.~Nie}, \bibinfo{author}{C.~Li}, \bibinfo{author}{L.~Tian}, \bibinfo{author}{Y.~Zhu}, \bibinfo{author}{Q.~Zhu}, \bibinfo{author}{G.~Dong}, \bibinfo{author}{J.~He}, \bibinfo{title}{{COVID-19 CT Lung and Infection Segmentation Dataset}}, \bibinfo{type}{Technical Report}, \bibinfo{year}{2020}. \URLprefix \url{https://doi.org/10.5281/zenodo.3757476}. \DOIprefix\doi{10.5281/zenodo.3757476}.
%Type = Article
\bibitem[{Buda et~al.(2019)Buda, Saha, and Mazurowski}]{buda2019association}
\bibinfo{author}{M.~Buda}, \bibinfo{author}{A.~Saha}, \bibinfo{author}{M.~A. Mazurowski},
\newblock \bibinfo{title}{Association of genomic subtypes of lower-grade gliomas with shape features automatically extracted by a deep learning algorithm},
\newblock \bibinfo{journal}{Computers in Biology and Medicine} \bibinfo{volume}{109} (\bibinfo{year}{2019}). \DOIprefix\doi{10.1016/j.compbiomed.2019.05.002}.
%Type = Inproceedings
\bibitem[{Zhu et~al.(2021)Zhu, Bashir, and Mazurowski}]{zhu2021deep}
\bibinfo{author}{Z.~Zhu}, \bibinfo{author}{M.~R. Bashir}, \bibinfo{author}{M.~A. Mazurowski},
\newblock \bibinfo{title}{Deep neural networks trained for segmentation are sensitive to brightness changes: preliminary results},
\newblock in: \bibinfo{booktitle}{Medical Imaging 2021: Computer-Aided Diagnosis}, volume \bibinfo{volume}{11597}, \bibinfo{organization}{International Society for Optics and Photonics}, \bibinfo{year}{2021}, p. \bibinfo{pages}{1159729}.
%Type = Article
\bibitem[{Chang(2000)}]{857802}
\bibinfo{author}{C.-I. Chang},
\newblock \bibinfo{title}{An information-theoretic approach to spectral variability, similarity, and discrimination for hyperspectral image analysis},
\newblock \bibinfo{journal}{IEEE Transactions on Information Theory} \bibinfo{volume}{46} (\bibinfo{year}{2000}) \bibinfo{pages}{1927--1932}. \DOIprefix\doi{10.1109/18.857802}.
%Type = Article
\bibitem[{Hong et~al.(2019)Hong, Yokoya, Chanussot, and Zhu}]{8528557}
\bibinfo{author}{D.~Hong}, \bibinfo{author}{N.~Yokoya}, \bibinfo{author}{J.~Chanussot}, \bibinfo{author}{X.~X. Zhu},
\newblock \bibinfo{title}{An augmented linear mixing model to address spectral variability for hyperspectral unmixing},
\newblock \bibinfo{journal}{IEEE Transactions on Image Processing} \bibinfo{volume}{28} (\bibinfo{year}{2019}) \bibinfo{pages}{1923--1938}. \DOIprefix\doi{10.1109/TIP.2018.2878958}.
%Type = Inproceedings
\bibitem[{Izmailov et~al.(2022)Izmailov, Kirichenko, Gruver, and Wilson}]{NEURIPS2022_fb64a552}
\bibinfo{author}{P.~Izmailov}, \bibinfo{author}{P.~Kirichenko}, \bibinfo{author}{N.~Gruver}, \bibinfo{author}{A.~G. Wilson},
\newblock \bibinfo{title}{On feature learning in the presence of spurious correlations},
\newblock in: \bibinfo{editor}{S.~Koyejo}, \bibinfo{editor}{S.~Mohamed}, \bibinfo{editor}{A.~Agarwal}, \bibinfo{editor}{D.~Belgrave}, \bibinfo{editor}{K.~Cho}, \bibinfo{editor}{A.~Oh} (Eds.), \bibinfo{booktitle}{Advances in Neural Information Processing Systems}, volume~\bibinfo{volume}{35}, \bibinfo{publisher}{Curran Associates, Inc.}, \bibinfo{year}{2022}, pp. \bibinfo{pages}{38516--38532}. \URLprefix \url{https://proceedings.neurips.cc/paper_files/paper/2022/file/fb64a552feda3d981dbe43527a80a07e-Paper-Conference.pdf}.

\end{thebibliography}
\end{document}